\documentclass[lettersize,journal]{IEEEtran}
\usepackage{amsmath,amsfonts}
\usepackage[ruled,vlined,linesnumbered]{algorithm2e}
\SetKwInOut{KwIn}{Input}
\SetKwInOut{KwOut}{Output}

\usepackage{array}
\usepackage[caption=false,font=normalsize,labelfont=sf,textfont=sf]{subfig}
\usepackage{textcomp}
\usepackage{stfloats}
\usepackage{url}
\usepackage{hyperref}
\usepackage{verbatim}
\usepackage{graphicx}
\usepackage{cite}
\usepackage{booktabs}
\begin{document}

\title{OPAL: Operator-Programmed Algorithms for Landscape-Aware Black-Box Optimization}

\author{Junbo~Jacob~Lian,
        Mingyang~Yu,
        Kaichen~Ouyang,
        Shengwei~Fu,
        Rui~Zhong,
        Yujun~Zhang,
        Jun~Zhang,~\IEEEmembership{Fellow,~IEEE}, and
        Huiling~Chen,~\IEEEmembership{Member,~IEEE}
\thanks{This research is financially supported by the National Natural Science Foundation of China (Grant No. 62076185, 62301367). \emph{(Corresponding authors: Jun Zhang \& Huiling Chen.)}}
\thanks{Junbo Jacob Lian is with the McCormick School of Engineering, Northwestern University, Evanston, IL, USA (e-mail: \texttt{Jacoblian@u.northwestern.edu}).}%
\thanks{Mingyang Yu is with the College of Artificial Intelligence, Nankai University, Tianjin, China (e-mail: \texttt{1120240312@mail.nankai.edu.cn}).}%
\thanks{Kaichen Ouyang is with the School of Mathematics, University of Science and Technology of China, Hefei, China (e-mail: \texttt{oykc@mail.ustc.edu.cn}).}%
\thanks{Shengwei Fu is with Guizhou University, Guiyang, China (e-mail: \texttt{gs.swfu22@gzu.edu.cn}).}%
\thanks{Yujun Zhang is with the School of New Energy, Jingchu University of Technology, Jingmen, China (e-mail: \texttt{zhangyj069@gmail.com}).}%
\thanks{Rui Zhong is with the Information Initiative Center, Hokkaido University, Sapporo, Japan (e-mail: \texttt{zhongrui@iic.hokudai.ac.jp}).}%
\thanks{Jun Zhang is with the College of Artificial Intelligence, Nankai University, Tianjin, China (e-mail: \texttt{junzhang@nankai.edu.cn}).}%
\thanks{Huiling Chen is with the School of Computer Science and Artificial Intelligence, Wenzhou University, Wenzhou, China (e-mail: \texttt{chenhuiling.jlu@gmail.com}).}%
\thanks{Source code, experiment scripts, and results are publicly available at \url{https://github.com/junbolian/OPAL}.}

}

\maketitle

\begin{abstract}
Black-box optimization often relies on evolutionary and swarm algorithms whose performance is highly problem dependent. We view an optimizer as a short program over a small vocabulary of search operators and learn this operator program separately for each problem instance. We instantiate this idea in Operator-Programmed Algorithms (OPAL), a landscape-aware framework for continuous black-box optimization that uses a small design budget with a standard differential evolution baseline to probe the landscape, builds a $k$-nearest neighbor graph over sampled points, and encodes this trajectory with a graph neural network. A meta-learner then maps the resulting representation to a phase-wise schedule of exploration, restart, and local search operators. On the CEC~2017 test suite, a single meta-trained OPAL policy is statistically competitive with state-of-the-art adaptive differential evolution variants and achieves significant improvements over simpler baselines under nonparametric tests. Ablation studies on CEC~2017 justify the choices for the design phase, the trajectory graph, and the operator-program representation, while the meta-components add only modest wall-clock overhead. Overall, the results indicate that operator-programmed, landscape-aware per-instance design is a practical way forward beyond ad hoc metaphor-based algorithms in black-box optimization.
\end{abstract}


\section{Introduction}
\IEEEPARstart{C}{ontinuous} black-box optimization (BBO) lies at the core of many engineering, control, and machine learning applications. In these settings, the objective function is expensive, derivative-free, noisy, or multimodal, and practitioners typically rely on population-based metaheuristics such as differential evolution (DE) and particle swarm optimization (PSO) \cite{Das2010DEsurvey,Pant2020DEreview,Kennedy1995PSO}. Over the last two decades, the field has produced a large ecosystem of increasingly sophisticated variants---for example L-SHADE \cite{Awad2016LSHADE} and jSO \cite{Brest2017jSO} in the DE family \cite{Zhong2025CFDE}---that achieve strong performance on standard benchmarks but remain highly problem-dependent. The No Free Lunch theorems for optimization make this dependence formal: averaged over all possible problems, no single algorithm or configuration can dominate all others \cite{Wolpert1997NFL}. In practice, even within a single problem instance, algorithmic needs change over time as the search progresses from global exploration to local exploitation \cite{Lian2025TAM}.

A natural response has been to use machine learning to support or replace human design. One line of work builds per-instance algorithm selectors and configurators: a feature extractor characterizes the problem, and a model predicts which algorithm or parameter setting will perform best \cite{Rice1976AlgSelection,Kerschke2019ELASelection}. Exploratory landscape analysis (ELA) plays a central role in this program: it maps sampled points into numerical descriptors of modality, ruggedness, conditioning, and other properties, which are then fed into supervised models for solver selection or performance prediction \cite{Kerschke2019ELASelection,Long2024LandscapeAAC}. Recent surveys document a rapid growth in such ML-assisted metaheuristics, covering both single-objective and multiobjective settings and highlighting hybrid designs where learned models steer classical metaheuristics \cite{Bolufe2025MLMetaheuristics}.

A second line of work treats the optimizer itself as a dynamical system and learns to adapt its parameters online. Automated and dynamic algorithm configuration (DAC) frameworks model configuration as a policy that maps a state representation of the search process to actions such as changing step sizes, mutation factors, or switching among update strategies \cite{Adriaensen2022DAC}. Benchmarks such as DACBench have standardized this view and enabled reinforcement-learning-based controllers to be compared across algorithms and problem families \cite{Eimer2021DACBench}. Subsequent work on instance selection for DAC shows how to choose training problems to improve cross-instance generalization of such policies \cite{Benjamins2024InstanceSelectionDAC}. In parallel, meta-black-box optimization and neural exploratory landscape analysis (NeurELA) push representation learning further, using deep networks to encode landscapes from samples in order to guide surrogate-assisted or model-based search \cite{Ma2024NeurELA,Du2025MetaBBO}. Landscape-aware automated configuration uses multi-output regression and classification on ELA features to map landscape fingerprints directly to algorithm parameters \cite{Long2024LandscapeAAC}.

Despite this progress, existing approaches share two structural limitations from the perspective of landscape-aware metaheuristic design. First, most methods operate either at the level of \emph{whole algorithms} or \emph{continuous parameter vectors}. Algorithm selection and configuration frameworks choose among a menu of complete solvers or tune a fixed set of hyperparameters \cite{Rice1976AlgSelection,Kerschke2019ELASelection,Adriaensen2022DAC,Long2024LandscapeAAC,Bolufe2025MLMetaheuristics}. Even dynamic configuration policies usually adjust scalar knobs of a given algorithm (e.g., $F$ and $CR$ for DE, or population size and inertia in PSO) rather than redesigning its internal logic \cite{Das2010DEsurvey,Pant2020DEreview,Adriaensen2022DAC}. At the other extreme, generative and program-synthesis approaches search directly in the space of operators or code: evolutionary design of evolutionary algorithms, explainable structure-generating DE, and program-synthesis tools such as GENESYS all demonstrate that algorithmic building blocks themselves can be learned or evolved \cite{Diosan2009EvoDesignEA,Gao2026SFGDE,Mandal2025GENESYS}. These methods can in principle discover novel heuristics, but they are costly to train, often fragile, and rarely provide per-instance adaptation at deployment time.

Second, there is a persistent bottleneck in how the state of the search is represented. Classical ELA typically computes global, hand-engineered features from an initial design of experiments; these ignore the specific trajectory followed by the optimizer and are expensive on high-dimensional problems \cite{Kerschke2019ELASelection,Bolufe2025MLMetaheuristics}. Dynamic configuration frameworks often collapse a rich population state into a small set of scalar indicators such as diversity, improvement rates, or success ratios \cite{Adriaensen2022DAC,Eimer2021DACBench,Benjamins2024InstanceSelectionDAC}. Neural landscape encoders such as NeurELA, landscape-aware DE and CMA-ES variants, and landscape-aware performance prediction models move towards richer representations, but still rely mainly on pooled statistics or problem-centric features rather than directly exploiting the structure of the sampled search trajectory \cite{Ma2024NeurELA,Lin2025LandscapeAwareDE,Wu2022LAMCMAES,Liefooghe2019LandscapeAwareEMO,Du2025MetaBBO}. As a result, the controller typically sees at best a blurred snapshot of the landscape, making it hard to take structured decisions about which search operators to deploy at which stage.

In this paper, we adopt a different perspective: an optimization algorithm is a \emph{short program over a vocabulary of search operators}, and the object we should learn \emph{per problem instance} is this operator program, not merely an algorithm label or a low-dimensional parameter vector. The operator vocabulary contains classic, well-understood building blocks from evolutionary computation and swarm intelligence---such as DE and PSO style variation operators, selection and replacement schemes, restart policies, and local search primitives---that can be shared across instances and tasks \cite{Das2010DEsurvey,Pant2020DEreview,Kennedy1995PSO,Awad2016LSHADE,Brest2017jSO}. Given a new black-box function, we aim to infer a phase-wise program that decides which operators to use during early exploration, mid-run refinement, and late-stage exploitation.

We formalize this view as the \emph{Landscape-to-Algorithm Operator Programs} (L2AO) problem. In L2AO, an optimizer first spends a small fraction of its evaluation budget in a \emph{design phase}, running a simple baseline algorithm to probe the landscape and collect an optimization trajectory. This trajectory is then encoded into a graph whose nodes are sampled solutions with local features (normalized fitness, distances, and improvement statistics), and whose edges connect nearby or sequential points. A graph neural network (GNN) processes this trajectory graph into a compact landscape embedding. Finally, a meta-learner maps the embedding to a short operator program: for each phase of the remaining budget it outputs a token from the operator vocabulary. Executing this program on the remaining evaluations yields a per-instance, landscape-aware algorithm that reuses known operators but recombines them in data-driven ways.

We present \emph{OPAL} (Operator-Programmed Algorithms), a concrete instantiation of L2AO for continuous single-objective black-box optimization. OPAL uses a parameter-fixed DE as a design-phase probe, builds a $k$-nearest neighbor trajectory graph, and applies a GNN-based policy to emit a three-phase operator program over a vocabulary including DE/PSO-style variation, restart, and axis-aligned local search. The meta-learner is trained once on a mixed distribution of synthetic tasks and a subset of CEC~2017 training functions covering unimodal, simple multimodal, hybrid, and composition landscapes, and is then applied out-of-distribution to the remaining CEC~2017 test functions without per-instance fine-tuning.

To assess OPAL, we conduct experiments on the CEC~2017 test suite in 10--100 dimensions. We compare against four strong baseline families: a classical DE with fixed parameters, a standard PSO variant, and two state-of-the-art adaptive DE algorithms, L-SHADE and jSO \cite{Das2010DEsurvey,Pant2020DEreview,Awad2016LSHADE,Brest2017jSO,Kennedy1995PSO}. On CEC~2017, OPAL attains competitive average ranks compared with state-of-the-art adaptive DE variants (L-SHADE and jSO), while significantly outperforming a classical DE and a standard PSO under Friedman and Holm post-hoc tests. We also quantify OPAL's meta-computation overhead, showing that graph construction, GNN encoding, and policy inference account for only a modest fraction of the total runtime compared with function evaluations.

Beyond aggregate performance, we perform ablation studies and qualitative analyses. Removing the trajectory-graph structure (OPAL-noGraph) and feeding only per-point features into the GNN significantly degrades results, underscoring the importance of explicit neighborhood information. Varying the meta-training distribution (OPAL-cecOnly) and turning off the auxiliary landscape-type head (OPAL-noAux) further shows that both mixed training tasks and explicit landscape supervision help stabilize the learned operator programs. The auxiliary head predicts coarse landscape categories (unimodal, simple multimodal, hybrid, composition) from the same graph embedding; this both improves robustness and reveals intuitive patterns in the learned programs. Analyzing the operator programs chosen by OPAL reveals intuitive behaviors---for example, increased use of restart and local search on rugged landscapes, and more conservative variation on smoother problems---which connect our method to existing landscape-aware designs \cite{Lin2025LandscapeAwareDE,Wu2022LAMCMAES,Liefooghe2019LandscapeAwareEMO,Long2024LandscapeAAC} while providing a unified, meta-learned framework.

In summary, the main contributions of this work are:
\begin{itemize}
  \item We introduce the \emph{Landscape-to-Algorithm Operator Programs} (L2AO) formulation, which casts per-instance algorithm design as learning short operator programs over a shared vocabulary of search operators, bridging algorithm selection, dynamic configuration, and generative hyper-heuristics.
  \item We propose \emph{OPAL}, a modular landscape-aware framework that (i) runs a short design phase using a simple baseline DE, (ii) builds a trajectory $k$-NN graph and embeds it with a GNN, and (iii) maps this embedding to a phase-wise operator program over DE/PSO-style, restart, and local search operators.
  \item We develop a practical meta-training procedure that combines REINFORCE with entropy regularization and an auxiliary landscape-type classification loss, and demonstrate on the CEC~2017 test suite that OPAL matches the performance of strong adaptive DE variants (L-SHADE and jSO) while significantly outperforming classical DE and PSO.
  \item Through ablation and overhead analyses on CEC~2017, we show that both the trajectory-based landscape embedding and the operator-program representation are crucial to OPAL's performance, while the additional computational cost of the meta-components remains modest relative to function evaluation time.
\end{itemize}
These results suggest that operator-programmed, landscape-aware per-instance design is a promising direction beyond the current proliferation of metaphor-based algorithms in black-box optimization, and provide a concrete blueprint for integrating landscape representations and operator libraries into future metaheuristic design.

The remainder of this article is organized as follows. Section~\ref{sec:related} reviews related work on classical and self-adaptive metaheuristics, algorithm selection, and automated algorithm design. Section~\ref{sec:opal-framework} presents the proposed OPAL model framework. Section~\ref{sec:experiments} reports experimental studies on the CEC~2017 test suite. Section~\ref{sec:conclusion} concludes the paper and outlines directions for future research.

\section{Related Work}\label{sec:related}

\subsection{Classical and Self-Adaptive Metaheuristics}
The DE family has spawned a large number of variants that differ in mutation strategies, parameter adaptation rules, and restart mechanisms, many of which have been carefully benchmarked on CEC test suites and engineering problems \cite{Das2010DEsurvey,Pant2020DEreview}. State-of-the-art schemes such as L-SHADE and jSO combine self-adaptation of control parameters with sophisticated population size reduction and archive mechanisms, and are widely used as strong baselines in the literature \cite{Awad2016LSHADE,Brest2017jSO}.

Beyond hand-designed variants, there is a long tradition of parameter control in evolutionary computation, including deterministic schedules, adaptive rules, and self-adaptive encodings of parameters into individuals \cite{Das2010DEsurvey,Pant2020DEreview}. Recent reviews emphasize that parameter dynamics can be as important as the choice of base operators themselves, and that many so-called ``new'' algorithms can be interpreted as particular parameter-control schemes layered on top of a small set of canonical templates \cite{Bolufe2025MLMetaheuristics}. However, in most cases, the structure of the algorithm---which operators exist and how they are composed---remains fixed, and adaptation operates in a relatively low-dimensional continuous parameter space.

Landscape-aware variants of classical algorithms represent a more targeted attempt to exploit problem structure. For example, landscape-aware DE designs mutation and selection strategies that react to estimated modality and basin structure \cite{Lin2025LandscapeAwareDE}, while cooperative coevolutionary CMA-ES uses landscape-aware grouping to decompose variables in noisy environments \cite{Wu2022LAMCMAES}. In multiobjective optimization, landscape-aware performance prediction models have been used to anticipate the behavior of evolutionary multiobjective optimizers based on features of the Pareto front and decision space \cite{Liefooghe2019LandscapeAwareEMO}. These methods support the idea that landscape information is valuable, but typically rely on hand-crafted diagnostics and remain tied to specific algorithm architectures.

OPAL builds on this line of work but moves the adaptation target from continuous parameters to discrete operator programs. Instead of manually designing problem-specific DE or CMA-ES variants, OPAL treats operators as reusable tokens and learns to assemble them into per-instance programs, guided by a learned representation of the sampled landscape.

\subsection{Algorithm Selection, ELA, and Dynamic Configuration}
Algorithm selection and automated configuration form the classical ML-based route to robust optimization performance. The algorithm selection problem, originally formulated by Rice \cite{Rice1976AlgSelection}, seeks to map problem features to a choice of solver. In continuous optimization, ELA provides a toolbox of numerical features extracted from sampled points---including measures of modality, curvature, and local search behavior---which have been successfully combined with machine-learning models for algorithm selection and performance prediction \cite{Kerschke2019ELASelection,Long2024LandscapeAAC}. Recent surveys highlight how such ML-assisted metaheuristics can systematically improve robustness across instance distributions, and review many designs where learned models drive or augment classical heuristics \cite{Bolufe2025MLMetaheuristics}.

However, classical algorithm selection and per-instance algorithm configuration (PIAC) typically choose a static configuration before the run starts. Dynamic algorithm configuration (DAC) generalizes this to online control: a policy observes a state representation of the current search process and outputs configuration actions such as adjusting parameters or mutation strategies \cite{Adriaensen2022DAC}. The DACBench library encapsulates a variety of DAC tasks for different optimizers and problem classes, making it possible to benchmark reinforcement-learning-based controllers that learn such policies \cite{Eimer2021DACBench}. Subsequent work addresses generalization and data efficiency issues, for example by selecting training instances to improve cross-instance transfer of DAC policies \cite{Benjamins2024InstanceSelectionDAC}.

Closely related are landscape-aware configuration and meta-BBO frameworks. Landscape-aware automated configuration has been studied using multi-output regression and classification to map ELA features to algorithm settings, effectively learning how to interpret landscape fingerprints in terms of hyperparameters \cite{Long2024LandscapeAAC}. Neural exploratory landscape analysis (NeurELA) replaces hand-crafted ELA features by learned representations, using deep networks to encode function samples and support algorithm selection and configuration \cite{Ma2024NeurELA}. Meta-black-box optimization with bi-space landscape analysis goes further by combining features from both the true objective and surrogate models within surrogate-assisted evolutionary algorithms, and learning dual-control mechanisms over the base optimizer \cite{Du2025MetaBBO}. These directions show how data-driven models can close the loop between measured landscape characteristics and algorithm behavior.

All of these approaches operate at the level of algorithm identities or continuous configuration vectors. They rarely manipulate the internal operator structure of algorithms, and their state representations are typically global or aggregated statistics (classical ELA, neural embeddings, or scalar performance indicators). In contrast, OPAL uses the sampled search trajectory itself as the primary state object, encoding it as a $k$-NN graph and processing it with a GNN to obtain a compact but structurally rich landscape embedding. The control object is also different: instead of adjusting a parameter vector, OPAL outputs a discrete, phase-wise operator program, which can be seen as a learned hyper-heuristic over a curated operator vocabulary.

\subsection{Automated Algorithm Design and Program Synthesis}
A more ambitious line of research tries to automatically design algorithms or operators themselves, rather than merely tuning existing ones. Evolutionary design of evolutionary algorithms uses genetic programming to evolve the structure of EAs, including selection, mutation, and recombination components, often targeting specific benchmark classes \cite{Diosan2009EvoDesignEA}. Recent work such as SFG-DE represents DE operators as compositional structures and evolves them, leading to explainable and adaptable DE variants that can continuously refine their operator graphs \cite{Gao2026SFGDE}. These approaches explore a rich design space of algorithmic building blocks, and can discover unconventional but effective operator structures.

In parallel, program-synthesis-based systems such as GENESYS treat algorithm design as synthesizing executable code \cite{Mandal2025GENESYS}. GENESYS couples evolutionary search with continuous optimization to evolve programs in a hybrid symbolic--numeric space, demonstrating that it is possible to learn non-trivial algorithms in numerical domains. Such systems are closely related to generative hyper-heuristics and neural program synthesis methods that attempt to learn search strategies end-to-end, and are surveyed in broader reviews of ML-enhanced metaheuristics \cite{Bolufe2025MLMetaheuristics}.

From a metaheuristic viewpoint, there is also growing interest in coupling classical solvers with machine learning to enhance individual components, for example by learning selection rules, adaptive neighborhood structures, or surrogate-based guidance \cite{Bolufe2025MLMetaheuristics,Lin2025LandscapeAwareDE,Wu2022LAMCMAES,Liefooghe2019LandscapeAwareEMO}. Landscape-aware DE and CMA-ES can be seen as hand-crafted special cases of such hybrids, where explicit landscape diagnostics drive algorithmic decisions \cite{Lin2025LandscapeAwareDE,Wu2022LAMCMAES}.

OPAL occupies an intermediate point between parameter control and full program synthesis. We do not attempt to invent new operators or arbitrary code; instead, we fix a curated vocabulary of interpretable operators drawn from DE, PSO, restart, and local search \cite{Das2010DEsurvey,Pant2020DEreview,Kennedy1995PSO,Awad2016LSHADE,Brest2017jSO}, and learn how to assemble them into short operator programs on a per-instance basis. This greatly reduces the search space compared with open-ended program synthesis, while still allowing rich algorithmic behavior to emerge from combinations of existing operators. By driving this assembly with a trajectory-graph-based landscape representation, OPAL ties automated algorithm design directly to what the optimizer actually sees during the run, rather than to static instance features alone, and connects naturally to the broader trends in landscape-aware configuration and meta-BBO \cite{Kerschke2019ELASelection,Long2024LandscapeAAC,Ma2024NeurELA,Du2025MetaBBO,Bolufe2025MLMetaheuristics}.

\section{OPAL Model Framework}
\label{sec:opal-framework}

\begin{figure}[t]
  \centering
  \includegraphics[width=\linewidth]{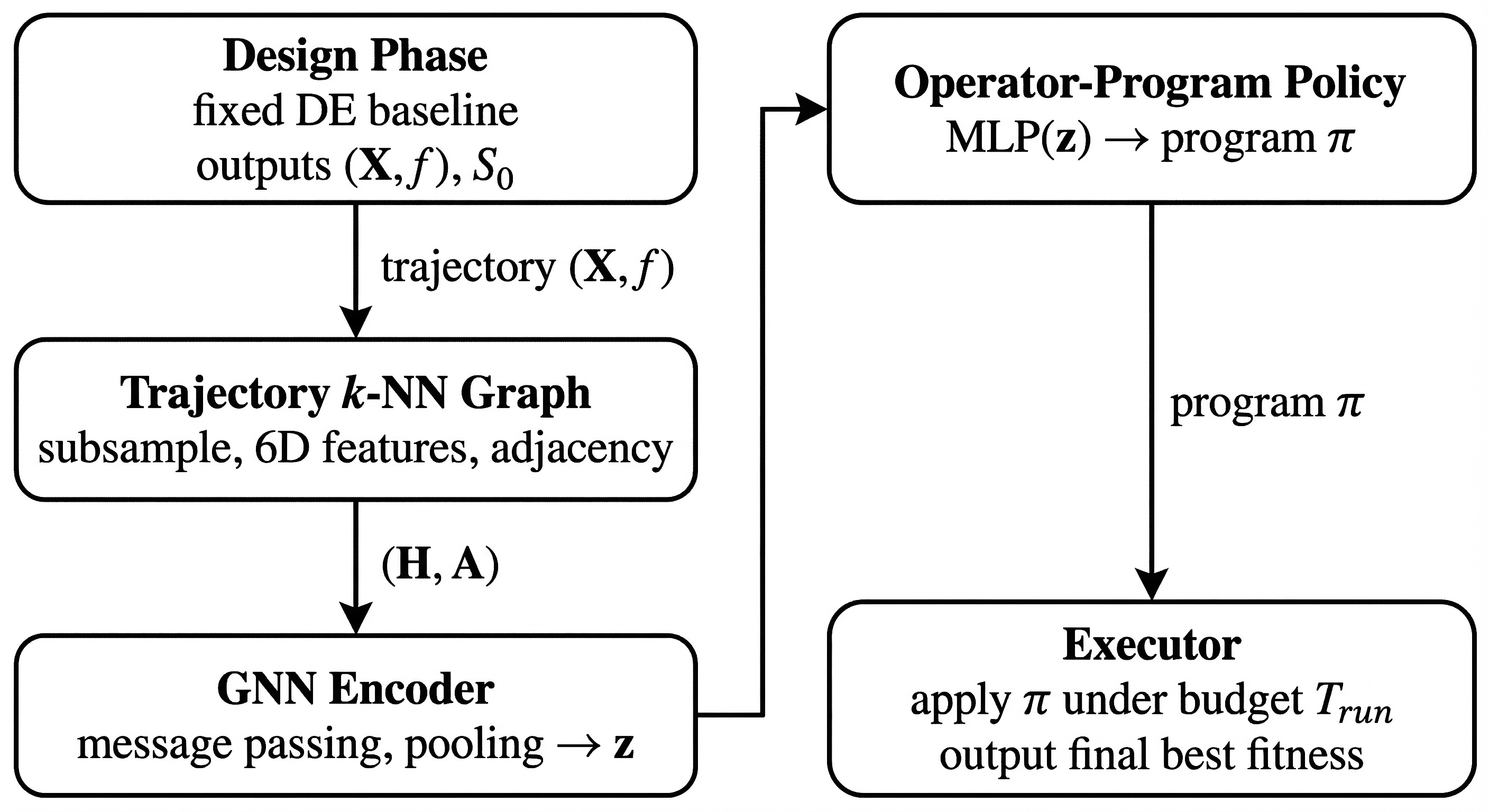}
  \caption{Overall OPAL workflow: a design phase with a fixed DE baseline,
  trajectory k-NN graph construction, GNN-based operator program generation,
  and execution of the learned program in the run phase. An auxiliary
  landscape-type classifier used during meta-training is omitted for clarity.}
  \label{fig:opal_flow}
\end{figure}

In this section we describe the proposed OPAL framework in detail. We first formalize the problem setting, then present the design-phase trajectory graph and its node features, the GNN-based policy that maps this graph to an operator program, the operator-program executor, the meta-training procedure, and finally a complexity analysis. The overall workflow is illustrated in Fig.~\ref{fig:opal_flow}.

\subsection{Problem Setting}

We consider continuous black-box minimization problems
\begin{equation}
    \min_{x \in \mathcal{X} \subset \mathbb{R}^d} f(x),
\end{equation}
where $f$ is expensive to evaluate, derivative-free, and possibly noisy or multimodal. The decision space $\mathcal{X}$ is a hyper-rectangle given by lower and upper bounds. A total budget of $T$ function evaluations (FEs) is available for each problem instance.

OPAL assumes that a simple baseline optimizer is available and treats this optimizer as a probe. A fraction $\rho \in (0,1)$ of the total budget is spent in a \emph{design phase} using this baseline to generate a trajectory of evaluated points. The remaining budget $(1-\rho)T$ is used by an operator-programmed algorithm instantiated by OPAL. In our experiments, we use a differential-evolution (DE) baseline with fixed parameters and a population of size $P$ \cite{Das2010DEsurvey,Pant2020DEreview}. The baseline follows a DE/rand/1/bin strategy with $F=0.7$, $CR=0.9$, and $P=50$.

Given the best objective value at the end of the design phase, $f_{\text{design}}^\star$, and the final best value after the operator program, $f_{\text{final}}^\star$, OPAL uses the logarithmic improvement
\begin{equation}
    R \;=\; \log_{10}\!\left(\frac{f_{\text{design}}^\star + \varepsilon}{f_{\text{final}}^\star + \varepsilon}\right),
\end{equation}
with a small $\varepsilon>0$, as reward for meta-training. This measure is approximately scale-invariant in $f$ and rewards both large improvements and robust gains across instances.

\subsection{Design-Phase Trajectory and k-NN Graph}

\subsubsection{Design-phase baseline}

In both meta-training and evaluation, OPAL calls a shared design-phase routine that runs the fixed DE variant for a given FE budget. Starting from a random population of size $P$ within the bounds of $\mathcal{X}$, the design phase performs standard DE mutation, crossover, and selection steps until it consumes
\[
    T_{\text{design}} \;=\; \lfloor \rho T \rfloor
\]
function evaluations. Every function evaluation produces a candidate $x \in \mathbb{R}^d$ and its objective value $f(x)$; these are stored as a trajectory. At the end of the design phase the routine returns:
\begin{itemize}
  \item the sequence of evaluated points $X \in \mathbb{R}^{M \times d}$,
  \item their objective values $f \in \mathbb{R}^M$,
  \item a population state $S_0$ containing the final population and its fitness.
\end{itemize}
Here $M$ denotes the total number of evaluated points in the design phase. In our meta-training implementation on synthetic and CEC-like tasks, we set the population size to $P=50$, the design ratio to $\rho=0.2$, and the total budget to $T_{\text{train}} = 1000 d$ function evaluations per episode. For CEC~2017 testbed experiments we adopt a larger budget $T_{\text{test}} = 10000 d$; see Section~4.

\subsubsection{Node features}

The design-phase trajectory encodes valuable information about the landscape. OPAL converts this trajectory into a graph structure. Given the matrix of evaluated points $X \in \mathbb{R}^{M \times d}$ and their scalar fitness vector $f \in \mathbb{R}^{M}$, we first select at most $M_{\max}$ points and then construct a $k$-nearest neighbor (k-NN) graph in the decision space. Each selected point becomes a node, and each node is annotated with a six-dimensional feature vector derived from the trajectory:
\begin{enumerate}
  \item \textbf{Z-scored fitness $f_z$:} the raw fitness values for the selected points are standardized to zero mean and unit variance,
  \[
      f_z(i) \;=\; \frac{f(i) - \mu_f}{\sigma_f + \varepsilon},
  \]
  where $\mu_f$ and $\sigma_f$ are the mean and standard deviation over the selected subset and $\varepsilon$ is a small constant.

  \item \textbf{Normalized rank $r_{\text{norm}}$:} we compute the rank of each selected point within this subset, where rank zero corresponds to the best fitness, and normalize by the maximum rank,
  \[
      r_{\text{norm}}(i) \;=\; \frac{\text{rank}(f(i))}{\max\{N-1,1\}},
  \]
  where $N$ is the number of selected nodes. This captures relative quality independent of scale.

  \item \textbf{Distance to best $d_{\text{best}}$:} let $x_{\text{best}}$ be the selected point with smallest $f$. For each node $i$, we compute the Euclidean distance
  \[
      d_{\text{best}}(i) \;=\; \bigl\|x_i - x_{\text{best}}\bigr\|_2,
  \]
  which measures how far a solution is from the current best in the decision space.

  \item \textbf{Normalized time index $t_{\text{norm}}$:} OPAL retains the original trajectory indices of the selected points. For a node corresponding to index $t \in \{0,\dots,M-1\}$, we define
  \[
      t_{\text{norm}}(i) \;=\; 
      \begin{cases}
        t / (M-1), & M > 1,\\[2pt]
        0, & M = 1,
      \end{cases}
  \]
  so that $t_{\text{norm}}$ ranges in $[0,1]$ and encodes when the point appeared during the design phase.

  \item \textbf{Local improvement signal $\Delta f_{\text{loc}}$:} to capture short-term progress along the trajectory, we sort the selected indices in ascending time order and, for each node $i$ that is not the earliest in this order, compute the fitness difference between its predecessor and itself,
  \[
      \Delta f_{\text{loc}}(i) \;=\; f(\text{prev}(i)) - f(i).
  \]
  Nodes that correspond to clear improvements have positive values. This signal is then standardized over all selected nodes to zero mean and unit variance.

  \item \textbf{Dimension feature $d_{\text{feat}}$:} we encode the problem dimension $d$ as a constant node feature
  \[
      d_{\text{feat}}(i) \;=\; \log(d+1),
  \]
  repeated for all nodes. This feature is not standardized across nodes and allows the encoder to condition on dimension while remaining invariant to the number of nodes.
\end{enumerate}
These features are assembled into a matrix
\[
    \mathbf{H} \in \mathbb{R}^{N \times 6}
\]
where each row corresponds to one selected point. In our experiments, we cap $N$ at $M_{\max}=300$.

\subsubsection{Subsampling and sampling strategies}

When the design-phase trajectory contains more than $M_{\max}$ points, OPAL subsamples a subset of indices $I \subset \{0,\dots,M-1\}$ with $|I|\leq M_{\max}$. We consider several sampling strategies:
\begin{itemize}
  \item \textbf{Time-uniform sampling:} selects indices that are approximately equally spaced along the trajectory, preserving temporal coverage.
  \item \textbf{Random sampling:} selects indices uniformly at random without replacement.
  \item \textbf{Fitness-stratified sampling:} partitions the fitness values into quantile bins and selects a similar number of points from each bin, encouraging coverage of different fitness levels.
  \item \textbf{Mixed sampling:} combines half time-uniform and half fitness-stratified samples, then deduplicates and, if needed, subsamples to $M_{\max}$. This is the default in our experiments.
\end{itemize}

\subsubsection{k-NN adjacency}

Given the selected points $X_{\text{sel}} \in \mathbb{R}^{N \times d}$, we construct a symmetric k-NN graph based on Euclidean distance in decision space. We compute the pairwise distance matrix
\[
    D_{ij} \;=\; \bigl\|x_i - x_j\bigr\|_2,
\]
then, for each node $i$, set directed edges to its $k_{\text{eff}} = \min\{k, N-1\}$ nearest neighbors, excluding itself. The resulting adjacency is symmetrized by taking the element-wise maximum with its transpose, and self-loops are added on the diagonal. The final adjacency matrix is
\[
    \mathbf{A} \in \mathbb{R}^{N \times N}
\]
with entries in $\{0,1\}$. We use $k=10$ and add self-loops in all experiments.

\subsection{GNN Encoder and Operator-Program Policy}

\subsubsection{Operator vocabulary and program representation}

In OPAL, an optimization algorithm is represented as a short program over a
fixed vocabulary of search operators. Each operator is identified by a name
and a small set of hyperparameters. An operator registry contains callable
implementations of classical variation, recombination, restart, and local
search components. Representative examples include:
\begin{itemize}
  \item DE-style mutation and binomial crossover operators,
  \item PSO-style global movement steps,
  \item Gaussian mutation operators acting on the population or the current best,
  \item restart operators that reinitialize a fraction of the worst individuals,
  \item axis-aligned local search around the current best solution.
\end{itemize}
We denote the set of operator names by $\mathcal{O}$ and its cardinality
by $|\mathcal{O}|$. In the implementation used in this paper, the registry
contains eight concrete operators:
\begin{itemize}
  \item \texttt{de\_rand\_1\_bin},
  \item \texttt{de\_best\_1\_bin},
  \item \texttt{uniform\_crossover\_pairs},
  \item \texttt{pso\_global\_step},
  \item \texttt{gaussian\_mutation\_self},
  \item \texttt{gaussian\_mutation\_best},
  \item \texttt{restart\_worst\_fraction},
  \item \texttt{local\_search\_best\_axis}.
\end{itemize}
An \emph{operator program} is a finite list of operator calls
\begin{equation}
    \pi \;=\; \bigl[(o_1,\theta_1),\dots,(o_{L_{\text{prog}}},\theta_{L_{\text{prog}}})\bigr],
\end{equation}
where $o_\ell \in \mathcal{O}$ and $\theta_\ell$ is a dictionary of
hyperparameters for step $\ell$. The executor (described in the next subsection)
applies these operators in sequence and, when requested, repeats the sequence
until it exhausts the available FE budget.

\subsubsection{Graph encoder}

Given the node-feature matrix $\mathbf{H} \in \mathbb{R}^{N \times 6}$ and adjacency matrix $\mathbf{A} \in \mathbb{R}^{N \times N}$ from the design phase, OPAL uses a graph neural network to produce a compact embedding of the landscape. The encoder consists of $L$ message-passing layers with shared weights and hidden dimension $h$:
\begin{equation}
    \mathbf{H}^{(0)} = \mathbf{H}, \qquad
    \mathbf{H}^{(\ell+1)} = \phi\!\left(\mathbf{A}, \mathbf{H}^{(\ell)}\right), \quad \ell = 0,\dots,L-1,
\end{equation}
where $\phi$ denotes a graph convolution-style update that aggregates information from neighbors and applies a learned affine transformation followed by a nonlinearity. In our implementation we use three layers ($L=3$) with hidden dimension $h=64$.

After the final layer, we apply a global pooling operator over nodes, for example mean pooling,
\begin{equation}
    z \;=\; \frac{1}{N}\sum_{i=1}^N \mathbf{H}^{(L)}_{i,:},
\end{equation}
to obtain a fixed-dimensional embedding $z \in \mathbb{R}^{h}$ that summarizes the sampled landscape and the design-phase trajectory.

\subsubsection{Policy over operator programs}

The GNN embedding $z$ is passed to a feedforward network that parameterizes a distribution over operator programs. We factor the program into a small number of segments that we refer to as phases. In the current implementation we use three phases and a single operator position per phase, so the program length is fixed to $L_{\text{prog}} = 3$. For each phase $p \in \{1,2,3\}$, a multilayer perceptron (MLP) maps $z$ to unnormalized scores over operator names:
\begin{equation}
    \ell_{p} \;=\; W_{p} z + b_{p}, \qquad
    \pi_{p}(o) \;=\; \frac{\exp(\ell_{p}^{(o)})}{\sum_{o' \in \mathcal{O}} \exp(\ell_{p}^{(o')})}.
\end{equation}
During meta-training, one operator name is sampled from each categorical distribution $\pi_p$, which produces a stochastic program of length three. At evaluation time, OPAL uses greedy selection by taking the operator with the largest probability in each phase. The sampled phases are concatenated to form the full program $\pi$.

This design allows the policy to represent nontrivial operator sequences while keeping inference cost negligible compared with function evaluations.

\subsubsection{Auxiliary landscape-type head}

During meta-training, we attach a lightweight auxiliary classifier to the same embedding $z$. This head predicts a coarse landscape label for the current task, such as unimodal, simple multimodal, hybrid, or composition, using a single hidden layer followed by a softmax over the corresponding classes. The landscape label is provided by the task generator: CEC~2017-based tasks inherit their official family label (unimodal, simple multimodal, hybrid, or composition), while analytic and neural-network-based landscapes are assigned labels by construction. The classifier incurs a cross-entropy loss $\mathcal{L}_{\text{aux}}$ and is discarded at test time. Empirically, this auxiliary objective stabilizes representation learning and encourages the GNN to encode information that correlates with landscape structure, which in turn leads to more robust operator programs.

\subsection{Operator-Program Executor}

Given an operator program and an initial population state from the design phase, OPAL executes the program through a generic operator-program executor. The executor treats the list of operator calls as a deterministic schedule and repeats it from the beginning when it reaches the end, until the FE budget allocated to the run phase is exhausted.

\subsubsection{Population state}

The population state includes the current decision vectors, their fitness values, and any algorithm-specific metadata used by operators (for example, velocity vectors for PSO-style steps or archives for certain DE variants). Each operator takes a population state and the environment as input, performs a number of function evaluations, and returns an updated state together with the number of evaluations used.

\subsubsection{Execution loop}

Algorithm~\ref{alg:opal-executor} summarizes the executor.

\begin{algorithm}[t]
\caption{OPAL operator-program executor}
\label{alg:opal-executor}
\KwData{Environment $E$; initial population state $S_0$; operator program $\pi = [(o_1,\theta_1),\dots,(o_{L_{\text{prog}}},\theta_{L_{\text{prog}}})]$; run-phase FE budget $T_{\text{run}}$; random seed $s$.}
\KwResult{Final state $S_{\text{final}}$; best-fitness trace $H_{\text{best}}$; FE trace $H_{\text{fe}}$.}
Initialize random generator $\mathrm{rng}$ with seed $s$\;
$S \gets$ copy of $S_0$\;
$T_{\text{used}} \gets 0$\;
Compute initial best fitness $f_{\text{best}} \gets \min S.\mathrm{fitness}$\;
Initialize histories $H_{\text{best}} \gets [f_{\text{best}}]$ and $H_{\text{fe}} \gets [0]$\;
\If{$L_{\text{prog}} = 0$}{
    \Return $(S, H_{\text{best}}, H_{\text{fe}})$\;
}
$i \gets 1$\;
\While{$T_{\text{used}} < T_{\text{run}}$}{
    $(o,\theta) \gets (o_i,\theta_i)$\;
    $i \gets i + 1$\;
    \If{$i > L_{\text{prog}}$}{
        $i \gets 1$\tcp*{wrap around}
    }
    Retrieve operator function $g$ for name $o$\;
    $(S, \Delta T) \gets g(S, E; \mathrm{rng}, \theta)$\;
    $T_{\text{used}} \gets $ $T_{\text{used}} + \Delta T$\;
    $f_{\text{best}} \gets \min S.\mathrm{fitness}$\;
    Append $f_{\text{best}}$ to $H_{\text{best}}$ and $T_{\text{used}}$ to $H_{\text{fe}}$\;
    \If{$T_{\text{used}} \geq T_{\text{run}}$}{
        \textbf{break}\;
    }
}
$S_{\text{final}} \gets S$\;
\Return $(S_{\text{final}}, H_{\text{best}}, H_{\text{fe}})$\;
\end{algorithm}

Each operator internally accounts for its own FE usage, so the total cost may slightly overshoot $T_{\text{run}}$. In our experiments this overshoot is negligible relative to the overall budget.

\subsection{Meta-Training Procedure}

OPAL is trained by episodic reinforcement learning over a distribution of synthetic and benchmark tasks. In each training episode, OPAL samples a task, runs the design phase, constructs the trajectory graph, samples a program from the policy, executes it, and updates the policy parameters based on the observed improvement. At evaluation time, OPAL uses the same design-phase and graph-construction procedures, but applies the policy in greedy mode to obtain a deterministic program, which is then executed without further training.

\subsubsection{Training task distribution}

The task generator constructs a mixture of training problems that includes:
\begin{itemize}
  \item continuous functions from the CEC~2017 benchmark, covering unimodal, simple multimodal, hybrid, and composition landscapes,
  \item classical analytic test functions such as Sphere, Rastrigin, Ackley, and Rosenbrock with random affine transformations and Gaussian noise,
  \item neural-network-based landscapes with random affine transformations and noise.
\end{itemize}
For each sampled task, the generator returns an environment object with bounds, a function-evaluation interface, the dimension $d$, and a total training budget $T_{\text{train}} = 1000 d$ function evaluations.

\subsubsection{REINFORCE with entropy and auxiliary losses}

Let $\theta$ denote the parameters of the GNN encoder and the operator-program policy. For each episode, OPAL:
\begin{enumerate}
  \item samples a task and constructs its environment $E$ with dimension $d$ and budget $T_{\text{train}}$;
  \item runs the design phase for $T_{\text{design}} = \lfloor \rho T_{\text{train}} \rfloor$ evaluations using the DE baseline, obtaining $(X, f)$, the initial state $S_0$, and the design-phase best $f_{\text{design}}^\star$;
  \item builds the trajectory graph $(\mathbf{H}, \mathbf{A})$;
  \item applies the GNN encoder and policy to $(\mathbf{H}, \mathbf{A})$ and samples an operator program $\pi$ together with its log probability $\log p_\theta(\pi)$ and the sum of per-decision entropies $H_{\text{ent}}(\pi)$;
  \item runs the executor with budget $T_{\text{run}} = T_{\text{train}} - T_{\text{design}}$ to obtain the final best fitness $f_{\text{final}}^\star$;
  \item computes the reward $R$ from $f_{\text{design}}^\star$ and $f_{\text{final}}^\star$;
  \item updates a scalar baseline $b$ by exponential smoothing and computes an advantage $A = R - b$;
  \item computes the auxiliary classification loss $\mathcal{L}_{\text{aux}}$ from the landscape-type head attached to $z$.
\end{enumerate}
The policy parameters are updated by minimizing the combined loss
\begin{equation}
    \mathcal{L}(\theta)
    \;=\;
    - A \cdot \log p_\theta(\pi)
    \;-\;
    \beta H_{\text{ent}}(\pi)
    \;+\;
    \lambda_{\text{aux}} \, \mathcal{L}_{\text{aux}},
\end{equation}
where the first term is the standard REINFORCE objective with advantage $A$, the second term is an entropy regularizer that encourages exploration (with coefficient $\beta > 0$), and the third term is an auxiliary classification loss weighted by $\lambda_{\text{aux}} \ge 0$. In our experiments we use a moving baseline with smoothing factor $\alpha \approx 0.9$, modest entropy regularization, and a small auxiliary-loss weight to avoid dominating the reinforcement signal. Gradients are clipped to a fixed norm to stabilize training.

Algorithm~\ref{alg:opal-training} summarizes the meta-training loop.

\begin{algorithm}[t]
\caption{Meta-training OPAL by REINFORCE with entropy and auxiliary losses}
\label{alg:opal-training}
\KwData{Task generator; design ratio $\rho$; total episodes $N_{\text{epi}}$; optimizer hyperparameters.}
\KwResult{Trained policy parameters $\theta^\star$.}
Initialize policy parameters $\theta$\;
Initialize baseline $b \gets 0$\;
\For{$e = 1$ \KwTo $N_{\text{epi}}$}{
    Sample task and environment $E$ with dimension $d$ and budget $T_{\text{train}} = 1000 d$\;
    $T_{\text{design}} \gets \lfloor \rho T_{\text{train}} \rfloor$, $T_{\text{run}} \gets T_{\text{train}} - T_{\text{design}}$\;
    Run design-phase DE on $E$ for $T_{\text{design}}$ FEs to obtain $(X,f)$, $S_0$, and $f_{\text{design}}^\star$\;
    Build $(\mathbf{H},\mathbf{A})$\;
    $(\pi, \log p, H_{\text{ent}}) \gets \text{GNN\_Policy}_\theta(\mathbf{H},\mathbf{A})$ with stochastic sampling\;
    Execute $\pi$ on $E$ with budget $T_{\text{run}}$ starting from $S_0$; record $f_{\text{final}}^\star$\;
    Compute reward $R$ from $f_{\text{design}}^\star$ and $f_{\text{final}}^\star$\;
    \eIf{$e=1$}{
        $b \gets R$\;
    }{
        $b \gets \alpha b + (1-\alpha) R$ with $\alpha \approx 0.9$\;
    }
    $A \gets R - b$\;
    Compute auxiliary classification loss $\mathcal{L}_{\text{aux}}$ from the landscape-type head\;
    $\mathcal{L} \gets - A \cdot \log p - \beta H_{\text{ent}} + \lambda_{\text{aux}} \mathcal{L}_{\text{aux}}$\;
    Take a gradient step on $\theta$ to minimize $\mathcal{L}$ with gradient clipping\;
}
\Return $\theta^\star$\;
\end{algorithm}

\subsection{Computational Complexity}

We briefly summarize the computational complexity of a single OPAL run on one problem instance. Let $d$ be the dimension, $T$ the total FE budget, $T_{\text{design}}=\rho T$ the design-phase budget, and $T_{\text{run}}=(1-\rho)T$ the run-phase budget.

\subsubsection{Design phase}

The design phase performs $T_{\text{design}}$ function evaluations. Each step involves constant-time vector operations in $\mathbb{R}^d$, so the additional overhead beyond the black-box evaluations is $O(T_{\text{design}} d)$ and is shared with any DE-style baseline.

\subsubsection{Trajectory graph construction}

Let $N \leq M_{\max}$ be the number of selected nodes. Subsampling indices is $O(M)$, where $M$ is the length of the raw trajectory. Computing the six node features is $O(Nd)$ for distances to the best and $O(N)$ for the remaining scalar features. Building the full pairwise distance matrix requires $O(N^2 d)$ time, followed by selecting nearest neighbors in $O(N^2)$ total time using partial selection. With $M_{\max}=300$ and $d \leq 100$ in our experiments, this cost is modest.

\subsubsection{GNN and policy inference}

Each message-passing layer performs a constant amount of work per edge. The symmetric k-NN graph has $O(kN)$ edges, so a GNN with $L$ layers and hidden dimension $h$ requires $O(L k N h)$ operations. For $L=3$, $k=10$, $N \leq 300$, and $h=64$, this cost is negligible compared with $T$ function evaluations. The subsequent MLP that maps the pooled embedding to operator logits is $O(|\mathcal{O}| L_{\text{prog}} h)$, where $L_{\text{prog}}$ is the program length, which is also small. The auxiliary classifier adds only $O(h C)$ work, where $C$ is the number of landscape classes.

\subsubsection{Run phase}

The executor calls operators until it consumes approximately $T_{\text{run}}$ evaluations. Each operator call performs $O(d)$ vector operations per evaluation, so the total overhead beyond function evaluations is $O(T_{\text{run}} d)$, comparable to classical population-based metaheuristics.

\subsubsection{Overall overhead}

Summing up, OPAL adds on top of a DE-style baseline:
\begin{itemize}
  \item one trajectory graph construction $O(N^2 d)$,
  \item one GNN and policy forward pass $O(L k N h)$ plus a small auxiliary head,
  \item bookkeeping within the executor.
\end{itemize}
For the parameter ranges used in our experiments, this overhead is well below the cost of function evaluations and results in a small constant-factor increase per run.

\section{Experimental Studies}
\label{sec:experiments}

We now evaluate OPAL on the CEC~2017 single-objective benchmark suite. We first describe the benchmark split, meta-training setup, baselines, and evaluation protocol. We then report aggregate results on CEC~2017, analyze function-level operator patterns, perform ablations on key components, and discuss limitations.

\subsection{Experimental Setup}
\label{subsec:exp-setup}

\subsubsection{Benchmarks and train--test split}

We adopt the 30-function CEC~2017 benchmark suite of shifted and rotated functions, excluding F2 as is common practice due to numerical issues. To study generalization across both \emph{landscape types} and \emph{dimensions}, we partition the remaining functions into a meta-training set and a disjoint test set.

The meta-training set
\[
    \mathcal{F}_{\mathrm{train}} = \{\text{F1, F4, F6, F8, F11, F14, F20, F23, F26, F29}\}
\]
covers unimodal, simple multimodal, hybrid, and composition landscapes. The test set
\[
    \mathcal{F}_{\mathrm{test}} = \{\text{Fi} : i \in \{1,\dots,30\},\, i \neq 2,\, \text{Fi} \notin \mathcal{F}_{\mathrm{train}}\}
\]
contains all remaining functions and is used only for final evaluation.

During meta-training, OPAL is trained on a mixture of tasks with dimensions
\[
    d \in \{10, 30, 50\},
\]
so that the policy sees problems at multiple scales without directly overfitting to the largest test dimension. During evaluation, we focus on
\[
    d \in \{30, 50, 100\},
\]
which are the most commonly reported CEC~2017 settings. The dimensions $d=30$ and $d=50$ lie inside the meta-training regime, while $d=100$ is treated as an out-of-distribution (OOD) dimension that is never seen during meta-training. All problems use the standard domain $[-100,100]^d$ with the official shifts and rotations.

To increase diversity beyond CEC, OPAL is also meta-trained on a small set of analytic functions (Sphere, Rastrigin, Ackley, Rosenbrock) under random affine transformations, as well as neural-network-based landscapes. These auxiliary tasks are not used for evaluation; they are only meant to regularize representation learning and reduce overfitting to a single benchmark family.

For each meta-training episode, the total evaluation budget is
\[
    T_{\mathrm{train}} = 1000 d,
\]
and a fixed fraction $\rho = 0.2$ is allocated to the design phase, as described in Section~\ref{sec:opal-framework}; the remaining evaluations are used to execute the sampled operator program.

\subsubsection{OPAL configuration}

During both meta-training and testing, the design phase runs a simple DE/rand/1/bin probe with population size $P=50$, mutation factor $F=0.7$, and crossover rate $CR=0.9$. For a given test budget $T$, the design phase consumes
\[
    T_{\mathrm{design}} = \lfloor \rho T \rfloor
\]
evaluations and the run phase receives
\[
    T_{\mathrm{run}} = T - T_{\mathrm{design}}.
\]

From the design-phase trajectory, OPAL builds a $k$-nearest neighbor graph with at most $M_{\max}=300$ nodes and $k=10$ neighbors, using the mixed subsampling strategy (half time-uniform, half fitness-stratified) and the six-dimensional node features defined in Section~\ref{sec:opal-framework}. The trajectory encoder is a three-layer message-passing GNN with hidden dimension 64 and mean pooling, and the policy head is a small MLP that outputs a three-phase operator program over the operator vocabulary. During meta-training, an auxiliary landscape-type classifier is attached to the same embedding; this head is discarded at test time.

The policy is optimized using REINFORCE with an advantage baseline, entropy regularization, and the auxiliary classification loss over $10{,}000$ episodes. Standard choices for learning rate, batch size, and gradient clipping are used and kept fixed across all experiments; full details are available in the released code.

\subsubsection{Baselines}

We compare OPAL against four baseline families that represent classical and state-of-the-art population-based optimizers:

\begin{itemize}
  \item \textbf{DE}: a classical DE/rand/1/bin with population size $P=50$ and fixed parameters $F=0.7$, $CR=0.9$~\cite{Das2010DEsurvey}. The algorithm runs for the full evaluation budget without restarts or self-adaptation.
  \item \textbf{PSO}: a global-best PSO~\cite{Kennedy1995PSO} with $P=50$ particles in a standard configuration (linearly decaying inertia weight and $c_1=c_2=2.0$), with velocities clamped to a fixed fraction of the search range.
  \item \textbf{L-SHADE}: the original L-SHADE variant with linear population size reduction and parameter memories, configured as in~\cite{Awad2016LSHADE}.
  \item \textbf{jSO}: the original jSO variant with hybrid parameter adaptation and archive management, configured as in~\cite{Brest2017jSO}.
\end{itemize}

All baselines operate directly on CEC~2017 objectives, use the same domain bounds and evaluation budgets as OPAL, and are run with their recommended default settings. None of them is meta-trained on $\mathcal{F}_{\mathrm{train}}$; they are applied in a standard “from-scratch” fashion to each test problem.

\subsubsection{Evaluation protocol and metrics}

For each $f \in \mathcal{F}_{\mathrm{test}}$, dimension $d \in \{30,50,100\}$, and algorithm, we run
\[
    N_{\mathrm{runs}} = 20
\]
independent trials with different random seeds. The total evaluation budget is
\[
    T_{\mathrm{test}} = 10000 d.
\]
For OPAL, the first $\lfloor 0.2 T_{\mathrm{test}} \rfloor$ evaluations are reserved for the design phase and the remaining evaluations for executing the learned operator program; the baselines use the full budget for their own update rules.

For each $(f,d)$ and algorithm, we record the best objective value at the end of the run, as well as the best-so-far trajectory as a function of the evaluation counter. We summarize performance via the mean, standard deviation, and median of the final best values over the 20 runs, and we compute average ranks across all test functions and dimensions. Statistical significance is assessed via Friedman tests on the rank matrix, with Holm-adjusted pairwise Wilcoxon signed-rank tests. We also report per-dimension Win/Tie/Loss (W/T/L) counts based on paired Wilcoxon tests at significance level $\alpha=0.05$.

\subsection{Comparison on CEC~2017}
\label{subsec:cec-results}

\begin{table}[t]
\centering
\setlength{\tabcolsep}{4pt}
\renewcommand{\arraystretch}{1.05}
\small
\caption{Overall statistical comparison on the CEC~2017 benchmark. Per-problem performance is measured by the median final objective value over 20 runs; smaller ranks are better.}
\label{tab:cec_overall}
\begin{tabular}{lccc}
\toprule
Algorithm & Avg. rank $\downarrow$ & Holm adj. $p$ (vs. ref.) & W/T/L (ref. vs. alg.) \\
\midrule
L-SHADE & 2.561 & 0.4452 & 19/20/18 \\
jSO      & 2.781 & 0.4452 & 17/11/29 \\
OPAL     & 2.798 & --     & --       \\
DE       & 3.140 & 0.2562 & 27/17/13 \\
PSO      & 3.719 & 3.6e-05$^{\dagger}$ & 44/4/9 \\
\midrule
\multicolumn{4}{l}{\footnotesize Ref. = OPAL. $^{\dagger}$ Holm-adjusted significance at $\alpha=0.05$.} \\
\bottomrule
\end{tabular}
\end{table}

\begin{table}[t]
\centering
\setlength{\tabcolsep}{3pt}
\renewcommand{\arraystretch}{1.05}
\scriptsize
\caption{Average rank by dimension on CEC~2017 (smaller is better).}
\label{tab:cec_by_dim}
\begin{tabular}{lccccc}
\toprule
Dim & OPAL & DE & PSO & L-SHADE & jSO \\
\midrule
30  & 2.763 & 3.053 & 3.737 & 2.368 & 3.079 \\
50  & 2.579 & 3.105 & 3.579 & 3.053 & 2.684 \\
100 & 3.053 & 3.263 & 3.842 & 2.263 & 2.579 \\
\bottomrule
\end{tabular}
\end{table}

\begin{table}[t]
\centering
\setlength{\tabcolsep}{3pt}
\renewcommand{\arraystretch}{1.05}
\scriptsize
\caption{Per-dimension Win/Tie/Loss (W/T/L) counts of OPAL against baselines on CEC~2017 (paired Wilcoxon tests, $\alpha=0.05$).}
\label{tab:cec_wtl_by_dim}
\begin{tabular}{lcccc}
\toprule
Dim & DE & PSO & L-SHADE & jSO \\
\midrule
30  & 9/6/4  & 15/1/3  & 5/9/5  & 6/6/7  \\
50  & 9/6/4  & 15/0/4  & 9/6/4  & 7/3/9  \\
100 & 9/5/5  & 14/3/2  & 5/5/9  & 4/2/13 \\
\bottomrule
\end{tabular}
\end{table}

We first summarize the overall performance across all CEC~2017 test functions at $d\in\{30,50,100\}$, and then break down the results by dimension. Tables~\ref{tab:cec_overall}, \ref{tab:cec_by_dim}, and~\ref{tab:cec_wtl_by_dim} report the corresponding statistics.

\paragraph{Overall performance.}

Table~\ref{tab:cec_overall} shows the global comparison on CEC~2017. Per-problem performance is computed using the median of final best values across 20 runs, and algorithms are ranked on each problem (smaller is better); the reported average rank is the mean of these per-problem ranks. The Friedman test on the rank matrix indicates statistically significant performance differences among algorithms ($\chi^2=18.673$, $p=0.0009$).

The three strongest methods (L-SHADE, jSO, and OPAL) form a tight cluster in terms of average rank (2.561, 2.781, and 2.798, respectively), and Holm-adjusted Wilcoxon tests do not detect significant differences between OPAL and either L-SHADE or jSO at $\alpha=0.05$. In other words, a single meta-trained OPAL policy is statistically \emph{competitive} with carefully hand-designed adaptive DE variants across the full CEC~2017 test suite.

At the same time, OPAL clearly improves over the simpler baselines. Compared with classical DE, OPAL wins substantially more problems than it loses (27/17/13 in W/T/L), and its average rank is lower (2.798 vs.\ 3.140). The gap to PSO is even more pronounced: OPAL achieves a much better average rank (3.719 for PSO), wins 44 problems and loses only 9, and the Holm-adjusted $p$-value for OPAL vs.\ PSO ($3.6\times 10^{-5}$) indicates a highly significant advantage.

\paragraph{Per-dimension behaviour.}

Table~\ref{tab:cec_by_dim} refines the picture by dimension $d\in\{30,50,100\}$. OPAL achieves its strongest relative performance at $d=50$ (average rank 2.579), where it is slightly ahead of jSO and close to L-SHADE. At $d=30$, OPAL remains competitive (2.763) but is marginally behind L-SHADE, which attains the best average rank in this easier regime. At $d=100$, OPAL’s average rank degrades to 3.053, while L-SHADE and jSO remain strong (2.263 and 2.579, respectively), indicating that the most optimized DE variants retain an edge in the hardest high-dimensional setting.

The W/T/L breakdown in Table~\ref{tab:cec_wtl_by_dim} makes this more concrete. Against PSO, OPAL consistently wins the majority of instances at all dimensions (e.g., 15/1/3 at $d=30$, 15/0/4 at $d=50$, and 14/3/2 at $d=100$), confirming a robust improvement over a widely used baseline. Against DE, OPAL is also favourable at every dimension (9/6/4, 9/6/4, and 9/5/5 from $d=30$ to $d=100$). By contrast, comparisons with L-SHADE and jSO are mixed: at $d=30$ and $d=50$, OPAL is roughly on par with L-SHADE and jSO (for example, OPAL vs.\ L-SHADE at $d=50$ yields 9/6/4, very close to parity), whereas at $d=100$ OPAL accumulates more losses, especially against jSO (4/2/13). This indicates that the performance gap with the strongest DE variants is largely a \emph{high-dimensional} phenomenon.

\paragraph{Analysis and discussion.}

The fact that OPAL is strongest at $d=50$ but degrades at $d=100$ is consistent with both the meta-training protocol and the algorithm design.

From a training-distribution viewpoint, OPAL is meta-trained only on $d\in\{10,30,50\}$. The 30- and 50-dimensional test problems therefore lie inside, or very close to, the training regime, while $d=100$ is an out-of-distribution dimension. In higher dimensions, the design-phase trajectories become sparser and the fixed-size $k$-NN graph ($M_{\max}=300$) covers a much thinner slice of the search space. This makes local neighborhoods less informative and reduces the effective resolution of the graph features fed to the GNN encoder, making it harder for the policy to accurately infer the landscape type and search stage.

From an algorithmic viewpoint, L-SHADE and jSO perform continuous, fine-grained adaptation of parameters and population size throughout the run. This kind of online adaptation remains very effective when the search dimension grows, because every generation can react to subtle changes in the population geometry. OPAL instead follows a “probe once, then commit to a program” strategy: the operator program is sampled after a single design phase and then executed without further meta-level adaptation. When the design-phase signal is degraded at $d=100$, committing to a single program can be suboptimal, and the lack of subsequent adaptation becomes a bottleneck. This explains why OPAL tracks L-SHADE and jSO closely at $d=30$ and $d=50$, but falls behind on some of the hardest 100-dimensional instances.

Overall, the CEC~2017 results show that a single meta-trained OPAL policy can match the performance of highly tuned adaptive DE variants (L-SHADE and jSO) on a broad set of benchmark functions, while clearly outperforming classical DE and PSO. At the same time, the observed degradation at $d=100$ points to a concrete avenue for future work: incorporating higher-dimensional tasks into the meta-training distribution and strengthening the design phase (e.g., dimension-aware node budgets or multi-stage design) to improve robustness in the most challenging high-dimensional regimes.

\subsubsection{Function-level patterns and operator programs}
\label{subsec:cec-function-patterns}

Beyond aggregate ranks, it is informative to examine how OPAL behaves on different CEC~2017 function families and how its learned operator programs adapt to the underlying landscape. The CEC~2017 suite groups functions into unimodal (F1--F3), simple multimodal (F4--F10), hybrid (F11--F20), and composition (F21--F30) classes, which roughly correspond to increasing structural complexity and interaction effects.

On the unimodal functions, all methods achieve very similar final errors across dimensions, and the per-problem Wilcoxon tests mostly report ties against both DE and the adaptive DE variants. This is expected: once a reasonably strong variation operator is in place, unimodal landscapes primarily test basic convergence speed rather than sophisticated adaptation, and OPAL's advantage from meta-learned operator programs is naturally limited in this regime.

On simple multimodal and most hybrid functions, OPAL still shows clear gains over the simpler baselines. For a large subset of these cases, the per-problem median errors of OPAL are lower than those of DE and PSO, and the Wilcoxon tests report wins rather than losses across both $d=30$ and $d=50$. This is consistent with the design-phase trajectory graph capturing multiple basins and funnel structures, so that the resulting landscape embedding allows OPAL to choose effective DE-style programs. However, the operator usage statistics reveal that, for unimodal, simple multimodal, and hybrid families, OPAL essentially converges to a \emph{single} default pattern dominated by DE operators: approximately two thirds of program phases use \texttt{de\_best\_1\_bin} and one third use \texttt{de\_rand\_1\_bin}, with virtually no use of auxiliary operators such as local search, Gaussian mutation, restart, or uniform crossover.

The picture becomes more nuanced on the hardest composition functions. At $d=100$, OPAL's median error is noticeably higher than that of L-SHADE and jSO on several rotated composition functions, even when it remains competitive at $d=30$ and $d=50$. In this regime, the fixed-size $k$-NN trajectory graph covers only a very thin slice of the high-dimensional search space; local neighborhoods become less informative, and the graph features fed to the GNN are less predictive of the global basin structure. Since OPAL commits to a single three-phase program after this design phase, a misleading initial trajectory can lock in a suboptimal operator sequence. By contrast, L-SHADE and jSO continuously adapt their parameters and population sizes throughout the run, allowing them to correct course even when early generations are not fully representative of the landscape.

\begin{figure}[t]
  \centering
  \includegraphics[width=\linewidth]{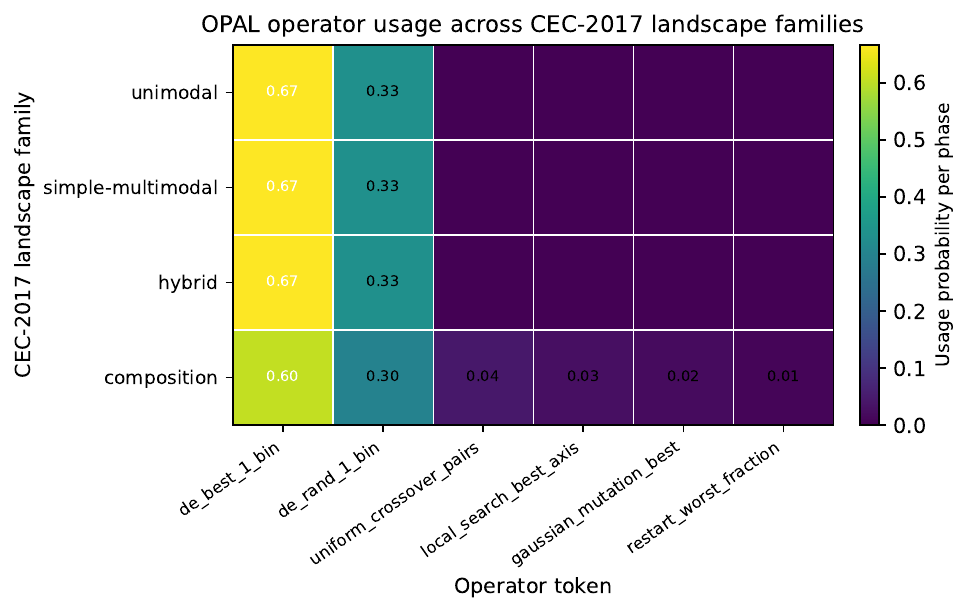}
  \caption{Operator usage frequencies of OPAL across CEC~2017 landscape families (aggregated over dimensions). Rows correspond to landscape families and columns to operator tokens; darker colors indicate higher usage probability. Only operators with nonzero usage on CEC~2017 are shown.}
  \label{fig:opal_operator_usage}
\end{figure}

To further probe OPAL's landscape-aware behavior, we log the operator programs emitted for each CEC~2017 test instance and group them by landscape family. Fig.~\ref{fig:opal_operator_usage} visualizes the empirical usage frequency of each \emph{used} operator token as a heatmap, with rows corresponding to the four CEC families and columns to the subset of the operator vocabulary that attains nonzero usage on CEC~2017. Two additional operators present in the registry (\texttt{pso\_global\_step} and \texttt{gaussian\_mutation\_self}) are never selected by the greedy policy on these benchmarks and therefore do not appear in the figure. Consistent with the discussion above, unimodal, simple multimodal, and hybrid functions are almost entirely served by the two DE-style tokens \texttt{de\_best\_1\_bin} and \texttt{de\_rand\_1\_bin}, with approximate probabilities $0.67$ and $0.33$, respectively. In other words, the meta-learner has discovered a robust “DE baseline program” that works well across these families, but does not strongly differentiate between them.

Composition functions are the only family where OPAL noticeably diversifies its programs. While they are still dominated by \texttt{de\_best\_1\_bin} and \texttt{de\_rand\_1\_bin} (together accounting for about $90\%$ of phases), roughly $10\%$ of phases use auxiliary tokens, including \texttt{uniform\_crossover\_pairs}, \texttt{local\_search\_best\_axis}, \texttt{gaussian\_mutation\_best}, and \texttt{restart\_worst\_fraction}. Inspecting the distinct operator programs reveals that these auxiliary operators appear in structured ways—for example, some programs start with restart or local-search tokens and then switch to DE-based exploitation with uniform crossover in the final phase. These patterns suggest that OPAL has learned to occasionally inject intensified local search or restarts on the most complex composition landscapes, but the overall probability of such programs remains relatively low.

Taken together, these observations support the interpretation of OPAL as a landscape-aware hyper-heuristic that reuses a curated vocabulary of classical operators but assembles them into instance-specific programs informed by the sampled trajectory. At the same time, the limited diversity of programs on hybrid and composition functions—especially in high dimensions—helps to explain why OPAL can closely match adaptive DE variants on many problems yet still lag behind them on some of the hardest 100-dimensional composition instances, where stronger or multi-stage program adaptation may be required.

\subsection{Ablation Studies}
\label{subsec:ablation}

To understand which components of OPAL are most influential, we conduct ablation studies on the CEC~2017 test suite at $d=50$ under the same evaluation budget and protocol as in Section~\ref{subsec:cec-results}. Across all variants, the design-phase DE probe, trajectory-graph construction (including $M_{\max}$, $k$, node features, and mixed subsampling), and the operator-program executor are kept fixed; only the meta-training configuration of the policy and its graph input are modified.

We evaluate four variants:

\paragraph{OPAL-full (main model).}
This is the default configuration described in Sections~\ref{sec:opal-framework} and~\ref{subsec:exp-setup}. The policy is meta-trained on the mixed task distribution (CEC~2017 train functions, affine-transformed analytic functions, and neural-network landscapes) at $d\in\{10,30,50\}$, using the full $k$-NN trajectory graph as input to the GNN encoder. The loss combines the REINFORCE objective with an advantage baseline, entropy regularization, and the auxiliary landscape-type classification loss with weight $\lambda_{\text{aux}} = 0.3$.

\paragraph{OPAL-noAux.}
This variant removes the auxiliary landscape-type supervision. The architecture is unchanged, but the auxiliary loss weight is set to $\lambda_{\text{aux}} = 0$, so the policy is trained purely from the improvement-based reinforcement signal plus entropy regularization. This variant isolates the role of explicit landscape labels in shaping the representation and the induced operator programs.

\paragraph{OPAL-cecOnly.}
This variant restricts the meta-training distribution to CEC~2017-based tasks only. The task generator draws problems exclusively from $\mathcal{F}_{\mathrm{train}}$ at $d\in\{10,30,50\}$ with the same per-episode budget and design ratio as OPAL-full, but no longer includes analytic or neural-network landscapes. The GNN architecture, auxiliary loss weight ($\lambda_{\text{aux}} = 0.3$), optimizer, and all other hyperparameters are identical to OPAL-full. This variant tests whether additional synthetic diversity beyond CEC is useful in practice.

\paragraph{OPAL-noGraph.}
This variant removes explicit neighborhood structure from the trajectory graph. During both meta-training and evaluation, the adjacency matrix fed to the GNN is replaced by the identity matrix, so message passing reduces to applying the same MLP independently to each node followed by global mean pooling. The node features, mixed training distribution, auxiliary loss weight, and optimization settings are the same as in OPAL-full. This variant isolates the effect of relational information between design-phase samples versus treating them as an unordered set.

All four variants are meta-trained for $10{,}000$ episodes with the same learning rate, optimizer, gradient-clipping threshold, and random-seed schedule. At test time, each variant uses greedy decoding of the three-phase operator program and is evaluated on the CEC~2017 test functions at $d=50$ with $N_{\mathrm{runs}} = 10$ independent trials under the same evaluation budget as in Section~\ref{subsec:cec-results}. For each function and variant, we record the median final objective value over the 10 runs and compute average ranks across the 19 CEC~2017 test functions. We also log the emitted operator programs to measure how often different operators are used.

Table~\ref{tab:opal_ablation_cec2017} summarizes the results. “Unique programs” counts distinct greedy operator programs emitted across all CEC~2017 test instances at $d=50$. “Non-DE frac.” is the fraction of operator tokens that are not \texttt{de\_rand\_1\_bin} or \texttt{de\_best\_1\_bin}.

\begin{table*}[t]
\centering
\setlength{\tabcolsep}{4pt}
\renewcommand{\arraystretch}{1.05}
\caption{Ablation results for OPAL variants on CEC~2017 at $d=50$.}
\label{tab:opal_ablation_cec2017}
\begin{tabular*}{\textwidth}{@{\extracolsep{\fill}}lcccccc}
\toprule
Variant & Train tasks & Graph & Aux. loss & Avg. rank $\downarrow$ & Unique programs & Non-DE frac. \\
\midrule
OPAL-full    & mixed      & k-NN     & yes & 2.368 & 4 & 0.025 \\
OPAL-noAux   & mixed      & k-NN     & no  & 2.579 & 1 & 0.667 \\
OPAL-cecOnly & CEC train  & k-NN     & yes & 2.053 & 1 & 0.000 \\
OPAL-noGraph & mixed      & identity & yes & 3.000 & 5 & 0.398 \\
\bottomrule
\end{tabular*}
\end{table*}

Several observations emerge. First, all four variants achieve broadly comparable average ranks on this moderate-sized $d=50$ test set. The advantage of OPAL-cecOnly over OPAL-full in Table~\ref{tab:opal_ablation_cec2017} is numerically small and does not exceed the Nemenyi critical distance for four algorithms over 19 problems at $\alpha=0.05$. We therefore do not treat OPAL-cecOnly as statistically superior to OPAL-full; the ablation is interpreted as diagnostic rather than as a separate benchmark competition.

Second, the diversity statistics show that OPAL-cecOnly reaches its slightly lower average rank by collapsing to a single operator program with purely DE-style tokens: across all 190 test runs at $d=50$ it produces one unique program and never uses non-DE operators (\texttt{non-DE frac.} $=0$). This behavior is consistent with strong tuning to the CEC training family and a loss of meaningful landscape differentiation. OPAL-full trades a small amount of CEC rank for more structured behavior, emitting four distinct programs and occasionally invoking auxiliary operators (about $2.5\%$ of tokens), mainly on harder functions. In line with the mixed-task design, OPAL-full is therefore used as the main model in the rest of the paper.

Third, removing the auxiliary landscape-type head (OPAL-noAux) degrades both performance and structure. OPAL-noAux attains a higher average rank (2.579) and collapses to a single program, but that program relies heavily on non-DE operators (\texttt{non-DE frac.} $\approx 0.67$) and does not translate this diversity into better objective values. The auxiliary head thus acts as a stabilizing regularizer on the representation and the resulting operator programs, rather than as a primary driver of raw CEC scores.

Finally, OPAL-noGraph, which replaces the $k$-NN adjacency by the identity matrix, exhibits the worst average rank (3.000) despite using the richest mix of operators (five distinct programs and nearly $40\%$ non-DE tokens). This confirms that the relational structure of the trajectory graph, not just the multiset of annotated samples, is important for producing informative landscape embeddings and effective operator-program choices.

Overall, the ablations indicate that OPAL's performance gains stem from the combination of (i) a mixed meta-training distribution that reduces overfitting to a single benchmark family, (ii) an auxiliary landscape-type objective that regularizes the encoder, and (iii) an explicit trajectory graph that exposes neighborhood structure to the GNN. The full configuration balances these ingredients, achieving competitive ranks on CEC~2017 while maintaining nontrivial yet controlled diversity in the learned operator programs.

\subsection{Limitations}
\label{subsec:limitations}

Our experimental design has several limitations that should be kept in mind when interpreting the results. First, the meta-training pool for OPAL is intentionally restricted to a subset of CEC~2017 functions, classical analytic benchmarks, and neural-network-based landscapes, with a total training budget of $T_{\text{train}} = 1000 d$ function evaluations per episode (Section~\ref{sec:opal-framework}). This choice keeps the CEC~2017 test functions strictly unseen and the benchmark protocol clean, but it also limits the diversity of landscapes that the policy is exposed to during training. A deployment-oriented variant of OPAL could be meta-trained on a broader and more application-driven task family, potentially including the full CEC~2017 suite and additional problem generators.

Second, the operator vocabulary used in the current implementation of OPAL is deliberately conservative. It contains a small set of standard DE- and PSO-style variation operators, simple restart rules, Gaussian mutations, and axis-aligned local search primitives (Section~\ref{sec:opal-framework}), but it does not expose many of the sophisticated mechanisms that make adaptive DE variants such as L-SHADE and jSO highly competitive on CEC benchmarks (for example, population-size reduction schedules, parameter memories, and archive-based selection) as separate operator tokens. As a consequence, OPAL consistently improves over classical DE and PSO, but it does not always surpass these heavily engineered adaptive DE baselines in our CEC~2017 experiments. Extending the operator library to factorize such mechanisms into reusable tokens, and re-training OPAL on this richer design space, is a natural extension of the present work.

Third, our empirical evaluation is limited to synthetic benchmarks, namely the CEC~2017 test suite. While this benchmark family is widely used and covers a variety of landscape types and dimensions, it does not include application-scale problems such as hyperparameter tuning, engineering design, or large-scale combinatorial optimization. Assessing OPAL on such real-world tasks is an important direction for future work.

Fourth, the ablation studies are conducted only in 50 dimensions, rather than across all four dimensions used in the full CEC~2017 comparison. In the main experiments, the relative ranking among OPAL and the baselines is broadly consistent across dimensions from ten to one hundred, and the ablation variants differ only in specific components while sharing the same underlying meta-training procedure and evaluation protocol. We therefore treat 50-dimensional problems as a representative mid-range setting that balances diagnostic value and computational cost, instead of repeating very similar ablations at 10, 30, and 100 dimensions.

Finally, our assessment of wall-clock overhead is based mainly on a complexity analysis and small-scale timing measurements. While these indicate that graph construction, GNN encoding, and policy inference contribute only a modest fraction of the total runtime compared to function evaluations, we do not provide a systematic profiling study across different problem classes, function-cost regimes, and hardware platforms. A more comprehensive empirical analysis of runtime, memory usage, and implementation-level overhead would be needed to fully characterize OPAL's computational footprint in large-scale applications.

\section{Conclusions}\label{sec:conclusion}

We have presented OPAL, a landscape-aware framework that treats a black-box optimizer as a short program over a small vocabulary of search operators and learns this operator program on a per-instance basis. OPAL uses a lightweight design phase with a fixed differential evolution probe to collect a trajectory of evaluated points, builds a $k$-nearest-neighbor graph over this trajectory, and encodes the resulting structure with a graph neural network. A meta-learner then maps the graph embedding to a phase-wise schedule of operators drawn from a curated library of DE-style steps, restarts, Gaussian mutations, and local search components. In this way, OPAL replaces hand-crafted metaphor-based algorithm design with data-driven, instance-specific operator programming.

On the single-objective CEC~2017 test suite, a single OPAL policy trained offline on a mixture of benchmark-based and synthetic tasks is statistically competitive with state-of-the-art adaptive DE variants such as L-SHADE and jSO, while clearly outperforming classical DE and PSO under the same evaluation budgets. Ablation studies further clarify which components are responsible for these gains: removing auxiliary landscape-type supervision or the trajectory-graph structure consistently degrades performance, and restricting training to CEC-only tasks yields slightly lower average ranks at the price of a nearly degenerate single DE-style program, pointing to overfitting to a narrow benchmark family rather than genuinely richer landscape adaptation. Across all settings, the overhead of graph construction, GNN encoding, and policy inference remains modest compared with the cost of function evaluations.

These findings suggest that operator-programmed, landscape-aware per-instance design is a practical path forward for continuous black-box optimization. Rather than proposing ever more specialized algorithms, one can expose a set of reusable operators, train a policy to assemble them into programs that respond to the observed trajectory, and reuse the resulting policy across problem instances. Future work includes enriching the operator vocabulary with factorized versions of advanced adaptive mechanisms, extending OPAL to multi-stage or online program adaptation in very high dimensions, and meta-training on broader, application-driven task families (including constrained, combinatorial, multiobjective, and real-world engineering problems). More generally, integrating operator programming with modern learning-based optimizers may help bridge the gap between principled algorithm design and the practical needs of large-scale black-box optimization in scientific and industrial settings.

\newpage

\begin{IEEEbiography}[ {\includegraphics[width=1in,height=1.3in,clip]{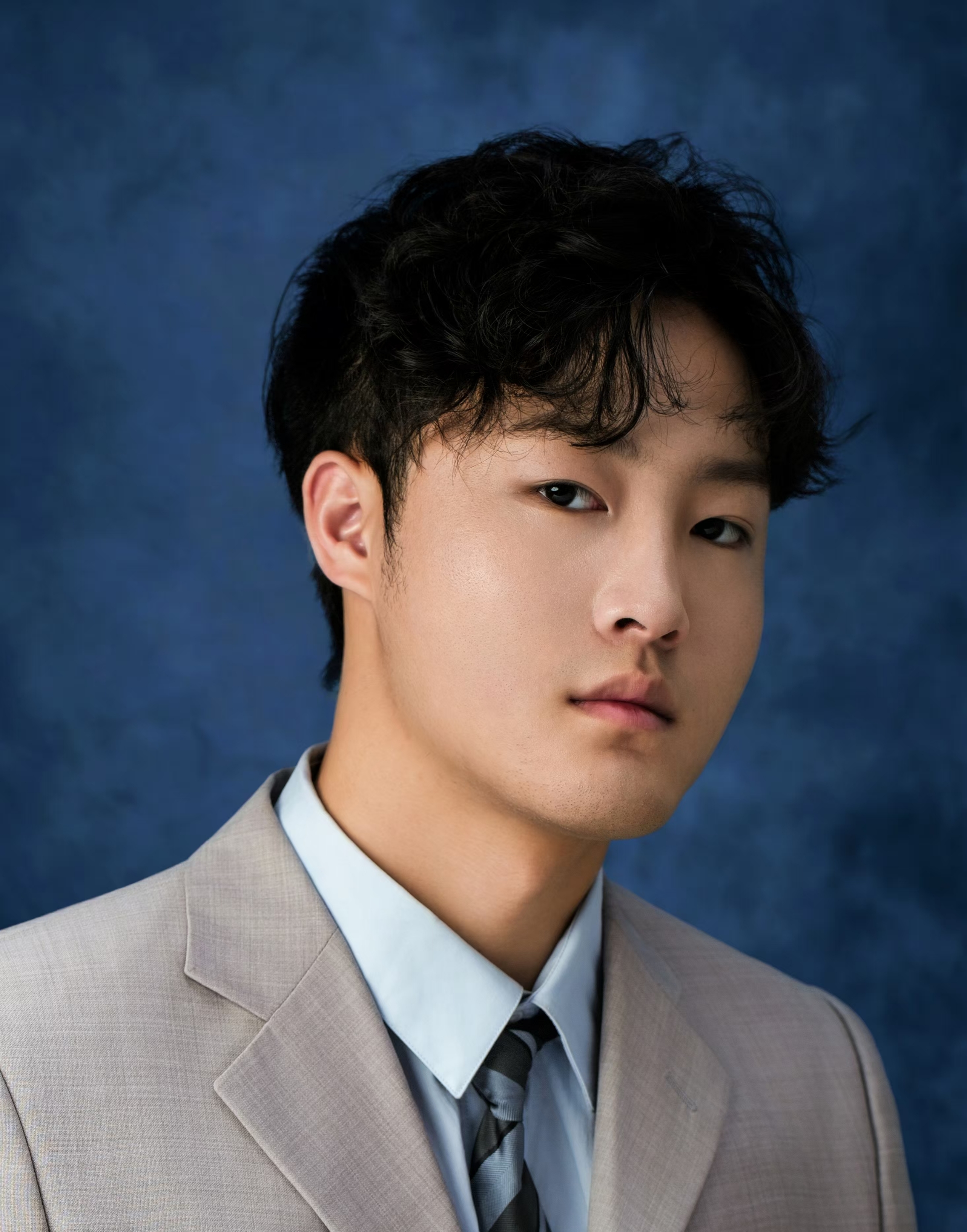}} ]{Junbo Jacob Lian} is currently pursuing his M.S. degree at Northwestern University, USA. His research interests include evolutionary computation, operations research, machine learning, and generative AI. He has published papers in international journals such as Neural Networks, Expert Systems with Applications, Swarm and Evolutionary Computation, Applied Soft Computing, Computers in Biology and Medicine, International Journal of Systems Science, and Measurement, and several of his works have been recognized as ESI highly cited and hot papers, including three ESI hot papers. He holds multiple national invention patents. He also serves as a reviewer for more than twenty international SCI journals. \end{IEEEbiography}

\begin{IEEEbiography}[
{\includegraphics[width=1in,height=1.3in,clip,trim={0 15 0 15}]{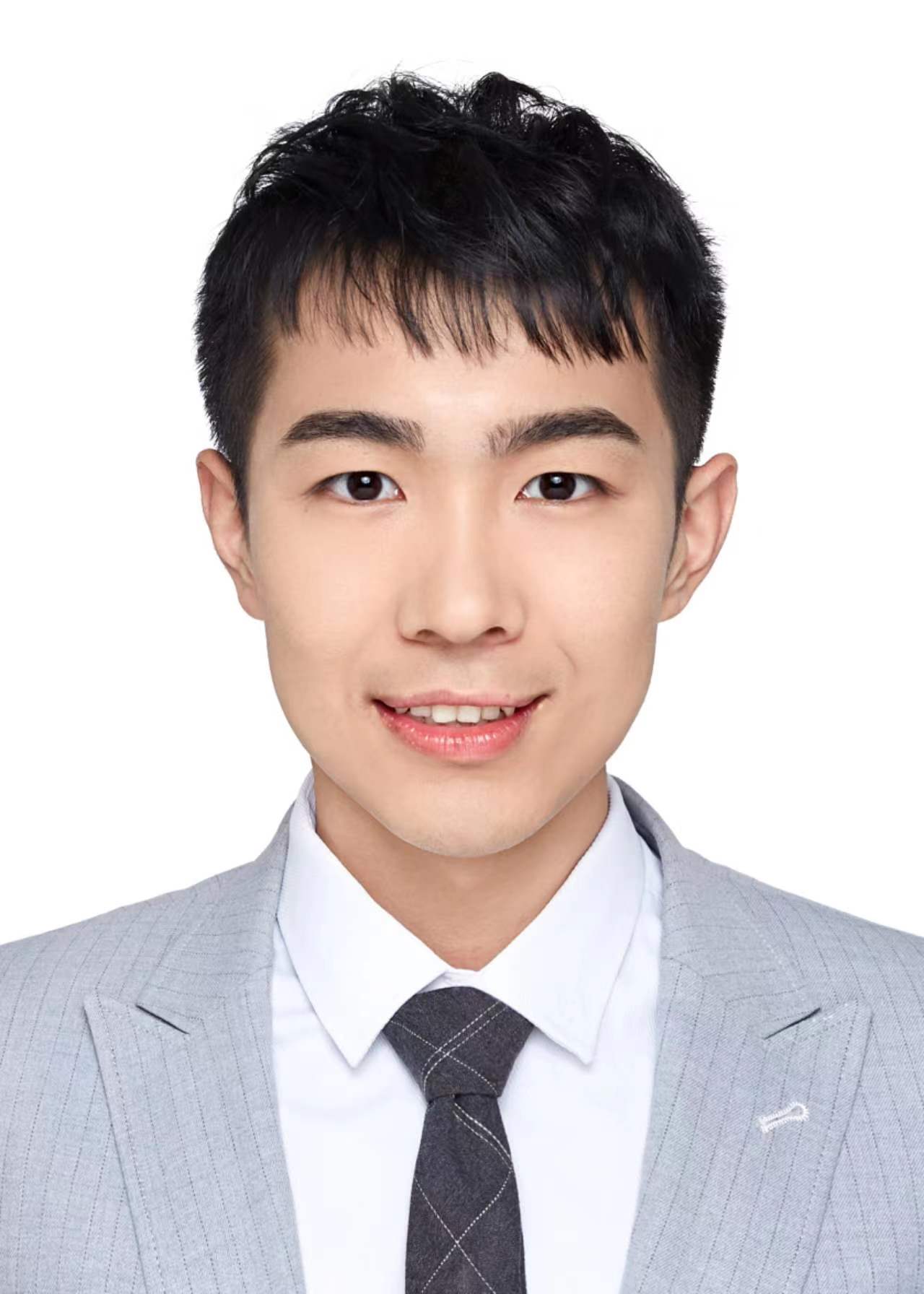}}
]{Mingyang Yu}
received his bachelor's degree from Jinan University's School of International Energy in 2022. Currently, he is pursuing a Ph.D. at the College of Artificial Intelligence, Nankai University. His research interests include swarm intelligence, evolutionary computation, and quantum computing, particularly their applications in edge computing and wireless sensor networks.\end{IEEEbiography}

\begin{IEEEbiography}[
{\includegraphics[width=1in,height=1.1in,clip,trim={0 35 0 35}]{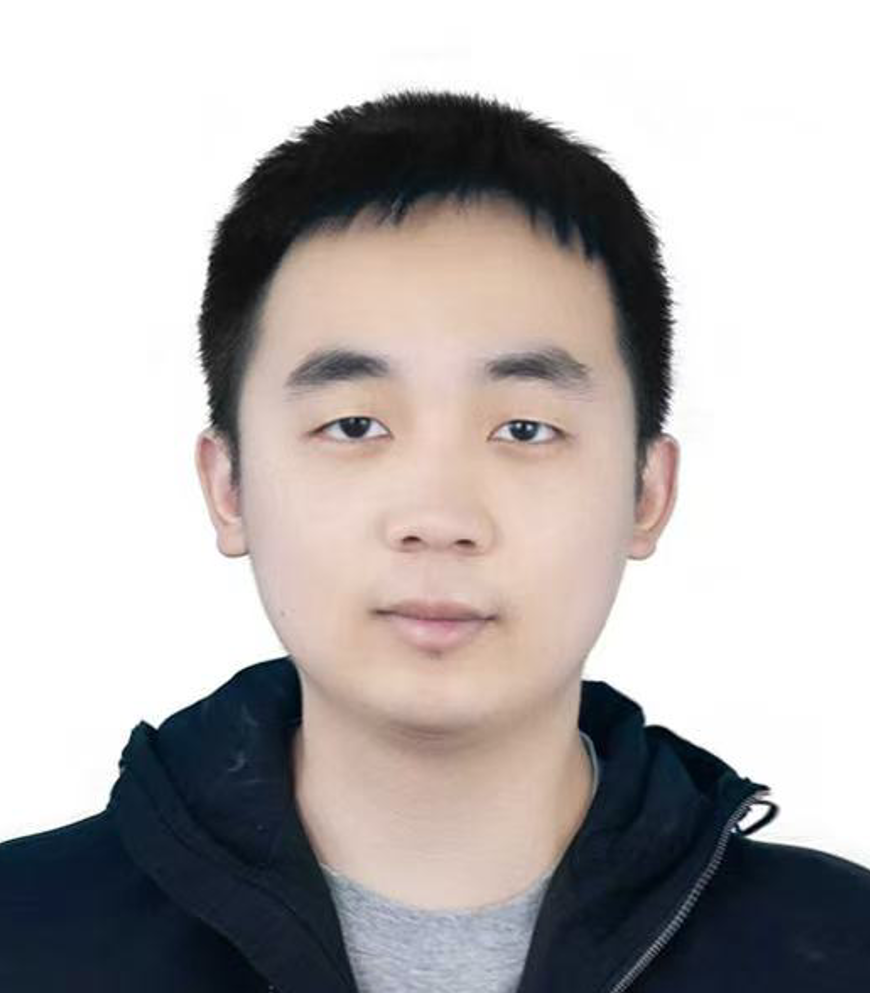}}
]{Kaichen Ouyang}
graduated from the University of Science and Technology of China with a Bachelor’s degree in Mathematics and Applied Mathematics. He is expected to pursue a Master’s degree in Applied Mathematics \& Statistics at Johns Hopkins University, USA. He serves as a reviewer for multiple JCR Q1 journals including Swarm and Evolutionary Computation, Knowledge-Based Systems,Information Sciences, Neurocomputing and Scientific Reports. His research interests include artificial intelligence, complex systems, mathematics and physics.
\end{IEEEbiography}

\begin{IEEEbiography}[
{\includegraphics[width=1in,height=1.3in,clip,trim={0 15 0 15}]{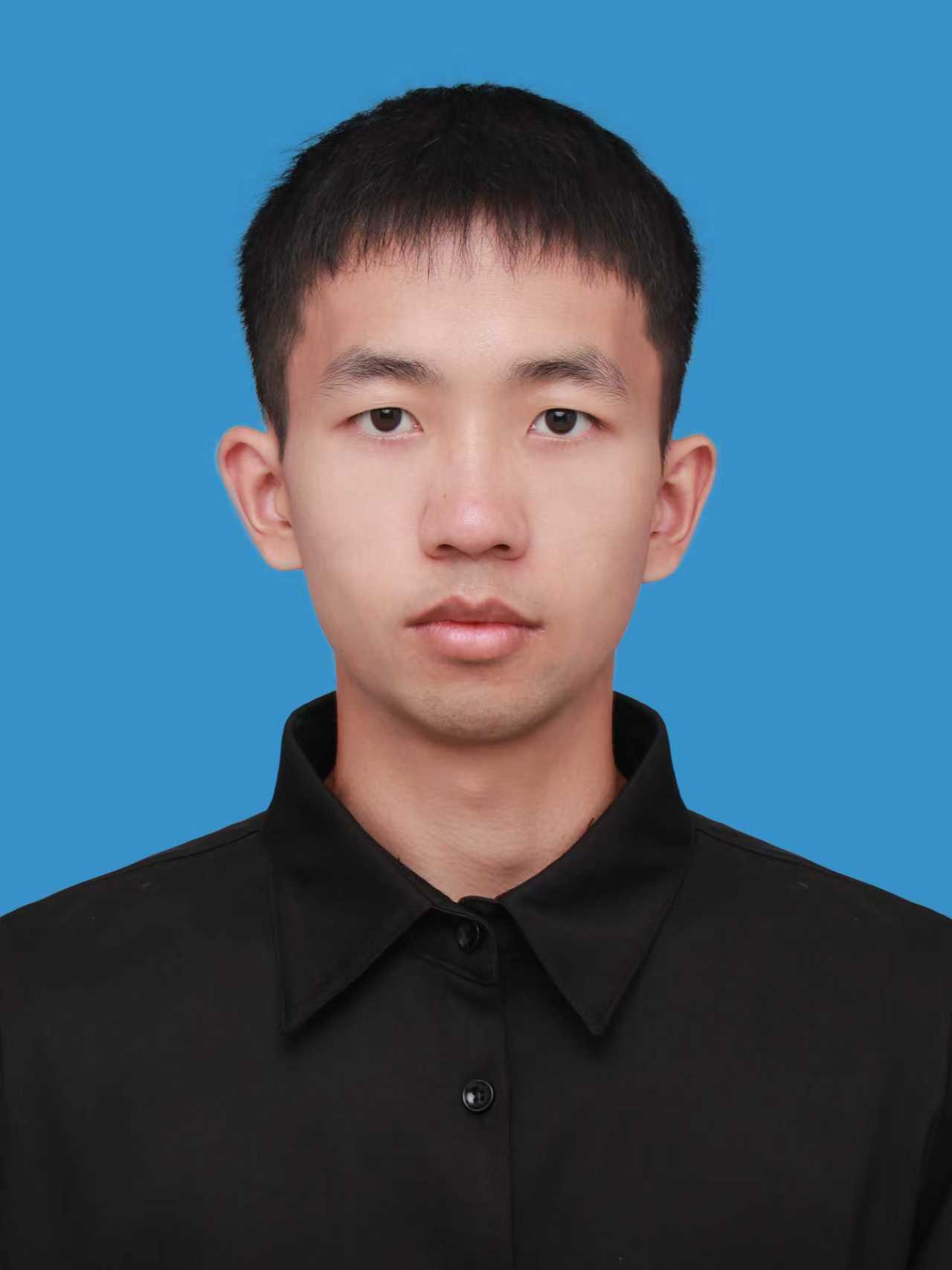}}
]{Shengwei Fu}
is a PhD candidate at Guizhou University.  And at the University of Auckland as a co-training PhD student. His research interests include multi-modal perception, human-robot collaboration,  and optimization algorithms. He has published papers and served as a reviewer in Artificial Intelligence Review,  Knowledge-Based Systems, Expert Systems with Applications, Applied Soft Computing, and others.
\end{IEEEbiography}

\begin{IEEEbiography}[
{\includegraphics[width=1in,height=1.2in,clip,trim={45 0 45 0}]{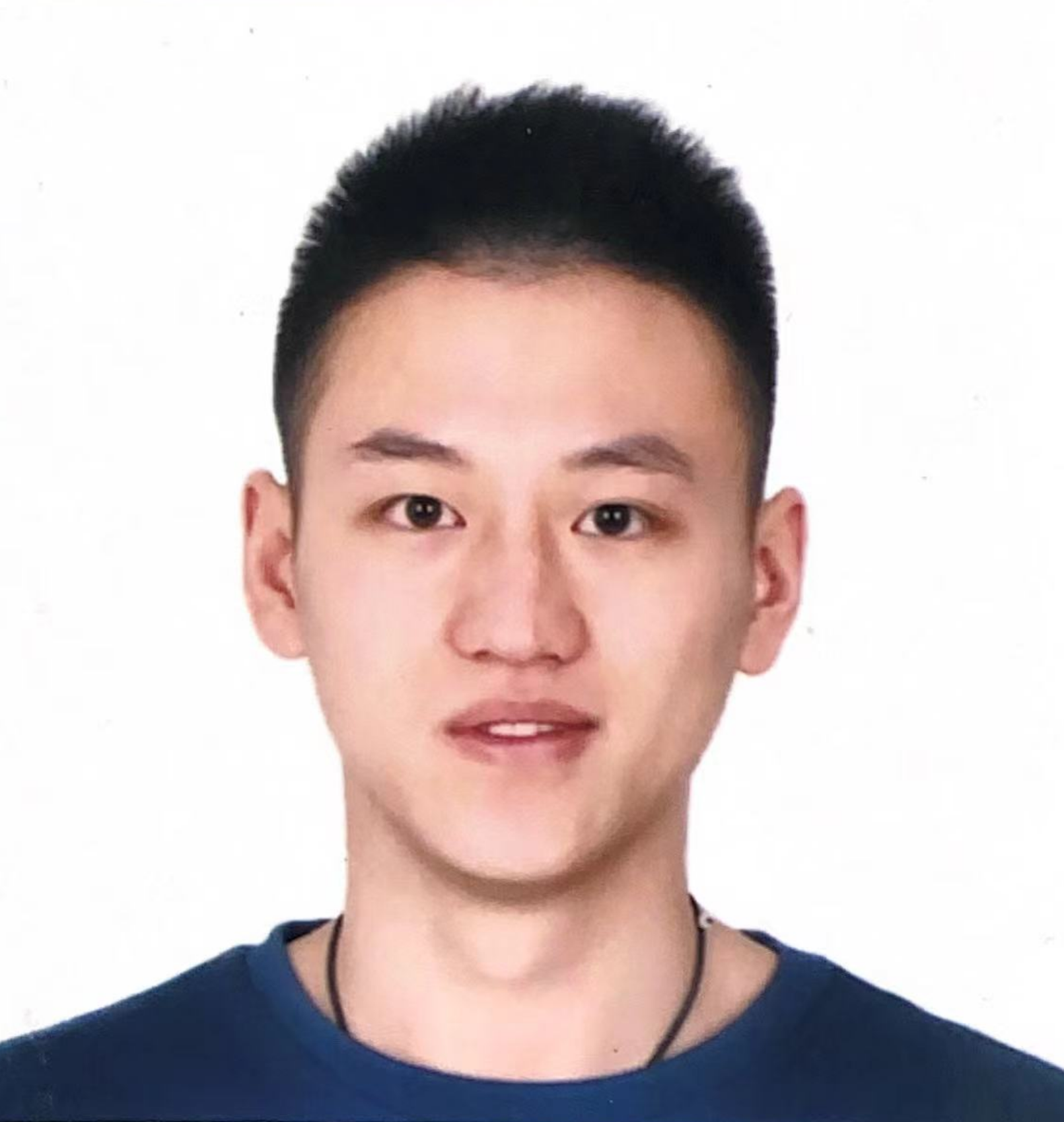}}
]{Rui Zhong}
received a bachelor’s degree from Huazhong Agricultural University, China, in 2019, a master’s degree from Kyushu University, Japan, in 2022, and a Ph.D. from Hokkaido University, Japan, in 2024. He is now a Specifically Appointed Assistant Professor at the Information Initiative Center, Hokkaido University. His research interests include Evolutionary Computation, Large-scale Global Optimization, and Meta-/Hyper-heuristics.
\end{IEEEbiography}

\begin{IEEEbiography}[
{\includegraphics[width=1in,height=1.3in,clip,trim={0 20 0 20}]{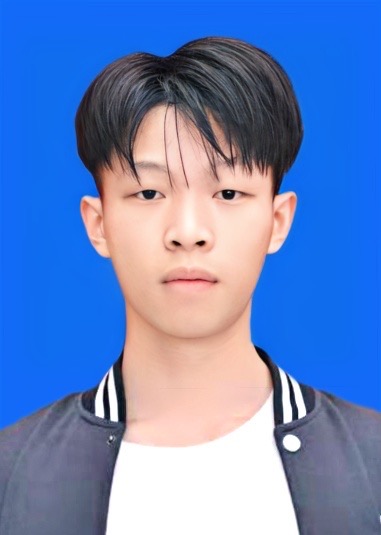}}
]{Yujun Zhang}
received his Bachelor of Engineering degree in 2023. He is pursuing a PhD in Computer Science with a research focus on evolutionary multi-task optimization and evolutionary many-task optimization. He has published several articles in journals such as Engineering Applications of Artificial Intelligence, IEEE ACCESS, and Alexandria Engineering Journal, etc. His research interests include swarm intelligence, evolutionary computation, evolutionary multi-task algorithms, evolutionary many-task algorithms, evolutionary multi-objective algorithms, nonlinear equation systems, photovoltaic model design, and UAV path planning.
\end{IEEEbiography}

\begin{IEEEbiography}[
{\includegraphics[width=1in,height=1.3in,clip,trim={0 15 0 15}]{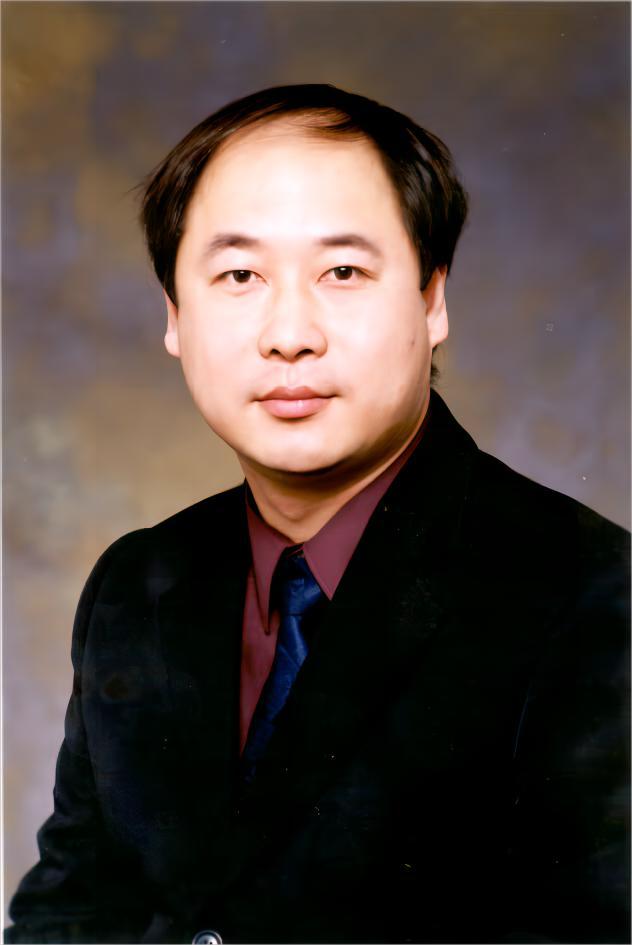}}
]{Jun Zhang}
(Fellow, IEEE) received the Ph.D. degree in Electrical Engineering from the City University of Hong Kong in 2002. He is currently with Nankai University, Hanyang University ERICA, Zhejiang Normal University, and Chaoyang University of Technology. His research interests include computational intelligence and evolutionary computation with applications in engineering. He has published more than 600 peer-reviewed papers, including over 230 in IEEE Transactions. Prof. Zhang received the National Science Fund for Distinguished Young Scholars in 2011 and was appointed Changjiang Chair Professor in 2013. He is a Clarivate Highly Cited Researcher and serves as an Associate Editor of the IEEE Transactions on Artificial Intelligence and the IEEE Transactions on Cybernetics.\end{IEEEbiography}

\begin{IEEEbiography}[
{\includegraphics[width=1in,height=1.3in,clip,trim={0 0 0 0}]{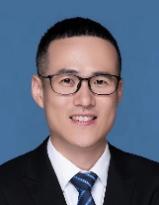}}
]{Huiling Chen}
 (Member, IEEE) is currently a professor in the college of computer science and artificial intelligence at Wenzhou University, China. He received his Ph.D. degree in the department of computer science and technology at Jilin University, China. His present research interests center on evolutionary computation, machine learning, data mining, and their applications to medical diagnosis, bankruptcy prediction, and parameter extraction of the solar cell. With more than 40000 citations and an H-index of 101, he is ranked worldwide among top scientists for Computer Science \& Electronics prepared by Guide2Research, the best portal for computer science research. He was served as the co-editor-in-chief of Computers in Biology and Medicine (2021.10–2024.12), and the editorial board member of Scientific Reports. He has published more than 200 papers in international journals and conference proceedings, including IEEE Transactions on Circuits and Systems for Video Technology, IEEE System Journal, IEEE Internet of Things Journal, Future Generation Computer Systems, Pattern Recognition, Expert Systems with Applications, Knowledge-Based Systems, and others. He has more than 40 ESI highly cited papers and 10 hot cited papers.
\end{IEEEbiography}

\vfill

\end{document}